\documentclass[11pt]{article}

\usepackage[preprint]{acl}
\usepackage{times}
\usepackage{latexsym}
\usepackage[T1]{fontenc}
\usepackage[utf8]{inputenc}
\usepackage{microtype}
\usepackage{inconsolata}
\usepackage{amsmath,amssymb}
\usepackage{booktabs}
\usepackage{graphicx}
\usepackage{adjustbox}
\usepackage{tabularx}
\usepackage{placeins}

\newcommand{\metrichead}[1]{\mbox{#1\,\ensuremath{\uparrow}}}

\title{Carving Radiology Reports at Their Evidential Joints:\\
Anatomical Region Sets as Generation Units}
\author{
  \textbf{Jinwei Yang, Junhao Chen, Jianwei Zhao,} \\
  \textbf{Zhonghan Xu, Hong Cheng, Ling Wang} \\
  University of Electronic Science and Technology of China
}

\hypersetup{
  pdftitle={Carving Radiology Reports at Their Evidential Joints: Anatomical Region Sets as Generation Units},
  pdfauthor={Jinwei Yang, Junhao Chen, Jianwei Zhao, Zhonghan Xu, Hong Cheng, Ling Wang}
}

\begin{document}
\maketitle

\begin{abstract}
Organizing radiology reports into independently generated units offers a way to generate fine-grained clinical content and bind it to visual evidence. However, existing methods typically adopt generation-unit structures not directly specified in natural reports or their annotations, introducing additional modeling compromises. We propose RadCarve, which instead defines generation units directly from the anatomical region sets in existing statement--region correspondences, unifying unit formation with visual evidence selection. RadCarve's planner uses composition-specialized query banks to jointly predict unit existence and region composition from the image, iteratively refining both with regional evidence aggregated according to current membership predictions. The generator independently produces each unit's text conditioned on the joint visual evidence of its predicted region set. Compared with strong baselines, RadCarve improves CXRFEScore by \textbf{9.4\%} in unit generation given region sets and achieves \textbf{\boldmath$2.4\times$} the end-to-end binding F1. It also achieves leading report-generation results on Chest ImaGenome and PadChest-GR, using fewer than one-sixth as many parameters as large-scale baselines.
\end{abstract}

\section{Introduction}
\label{sec:intro}

Automatically generating radiology reports from medical images, such as chest X-rays, could improve reporting efficiency and reduce radiologists' workload \citep{jing2018automatic,huang2025efficiency}. Compared with general vision--language tasks, radiology report generation has a lower tolerance for errors: models must identify the anatomical locations of findings and accurately describe the corresponding clinical observations \citep{yu2023evaluating}, because errors in a report can have serious clinical consequences \citep{bannur2024maira2}. Decomposing a report into independently generated units offers a way to organize content and associate it with evidence at a finer granularity. Recent research further shows that independent unit-level generation yields substantial performance gains \citep{cheng2026rethinking}. This raises a central modeling question: what should constitute a generation unit?

We argue that a key challenge for existing unit-based generation methods is that their adopted generation-unit structures are not directly specified in natural reports or their annotations, introducing additional modeling compromises. Methods with predefined topic schemes require report rewriting to construct aligned supervision; at inference, the units to generate can be specified by fixed topic lists or even ground-truth disease labels \citep{wang2022ita,cheng2026rethinking}. Region-level methods treat individual anatomical regions as generation units, fragmenting the joint visual evidence for statements involving multiple regions \citep{tanida2023rgrg}. \citet{dallaserra2023controllable} merge overlapping region sets together with their associated statements, yielding coarser region-set--text supervision.

Natural reports and their annotations, however, directly provide a statement--region structure: each statement $s_j$ is associated with an anatomical region set $A_j$. These correspondences are set-valued and overlapping: a statement may be supported by multiple regions, and a region may support multiple statements. In the Chest ImaGenome training data \citep{wu2021chest}, 85.2\% of the region sets contain two or more regions (Table~\ref{tab:pruning_impact}).

Building on these correspondences, we propose RadCarve, which follows the generation process $I \rightarrow R_t \rightarrow y_t$: given an image $I$, each predicted region set $R_t$ conditions independent generation of its unit text $y_t$, and the unit texts form a complete report. This design addresses this challenge in two ways. First, generation units are defined directly from the region sets in existing statement--region correspondences, without topic-based report rewriting, restricting generation units to individual regions, or merging region sets through shared regions (Table~\ref{tab:unit_construction}). Second, a region-set planner jointly predicts from the image which units exist and which regions each comprises, rather than relying on units supplied in advance. Each region set thus defines both a generation unit and its visual evidence, making the generated text traceable to the selected regions.

Our main contributions are:
\begin{itemize}
  \item \textbf{Defining generation units directly from existing statement--region correspondences.}
  We take the anatomical region sets associated with statements as generation units, unifying unit formation with visual evidence selection.
  \item \textbf{RadCarve, a region-set planning and generation framework.}
  We develop composition-specialized query banks and iterative composition refinement to jointly predict unit existence and region composition from images, supporting independent text generation from the selected regional evidence.
  \item \textbf{Validating the effectiveness of region-set generation.}
  Experiments show that RadCarve recovers multi-region bindings more accurately and that joint regional evidence supports more complete clinical content. RadCarve achieves leading in-domain report-generation results on Chest ImaGenome and the highest scores on all three clinical metrics in external evaluation on PadChest-GR.
\end{itemize}

\section{Related Work}
\label{sec:related}

\subsection{Spatial Grounding and Local Evidence in Complete Report Generation}
\label{sec:rw_grounding}

Prior work incorporates spatial information through class activation maps \citep{wang2024camanet,bu2024ekagen,wang2025ammrg}, organ segmentation masks \citep{nimalsiri2023mergis,gu2024comg,gu2025orid,sharma2025mairaseg}, anatomical scene graphs \citep{zhang2024apisg,wang2025sgrrg}, and detection-based regional prompts \citep{gao2026anatomyguided,hashmi2026anatomix}. Another line of work makes localization part of the report output: MAIRA-2 generates radiological findings with associated bounding boxes \citep{bannur2024maira2}; CURE's grounded report generation mode follows this setting and improves text--location alignment \citep{messina2026cure}. These approaches tie generated text more closely to spatial evidence but still produce the complete report through a single continuous autoregressive decoding process. In controlled experiments, \citet{cheng2026rethinking} show that the performance gap between real and blank images narrows as more preceding report text is supplied, indicating that later content increasingly relies on textual context rather than the image. Thus, providing or predicting spatial information does not guarantee that each statement's content is determined by its corresponding visual evidence.

\subsection{Generation Units in Radiology Report Generation}
\label{sec:rw_units}

Unit-based methods differ in what constitutes a generation unit and who specifies it. VTI, TranSQ, Multi-Grained RRG, and Teaser learn latent topics or semantic intentions using natural sentences as supervision proxies \citep{najdenkoska2022vti,kong2022transq,liu2024multigrained,zhao2024teaser}. Sentence boundaries, however, need not coincide with topic boundaries. ITA and LLaVA-TA instead group or rewrite reports to match predefined unit schemes; the units to generate at inference are externally determined \citep{wang2022ita,cheng2026rethinking}.

\begin{figure*}[t]
  \centering
  \includegraphics[width=\textwidth]{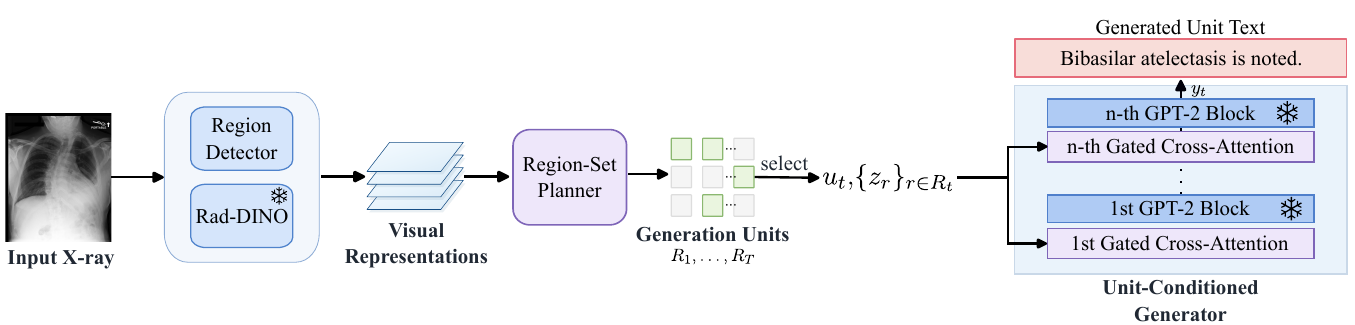}
  \caption{Overview of RadCarve. The planner predicts region sets $R_t$, each defining a generation unit and its visual evidence. The generator independently produces unit texts $y_t$, which are combined into a complete report. Snowflakes indicate frozen components.}
  \label{fig:overview}
\end{figure*}

Anatomy-based methods provide explicit spatial units. RGRG selects regions from the image and generates text independently for each, projecting the correspondences of cross-region statements onto single-region supervision and conditioning each unit on only one region's visual features \citep{tanida2023rgrg}. AOR instead splits cross-region sentences into single-region statements \citep{li2025aor}. However, in the manually annotated PadChest-GR, even a sentence containing a single radiological finding can correspond to multiple regions \citep{castro2025padchestgr}. In vision--language pre-training, ASG reports better downstream performance when aligning sentences with merged anatomical boxes rather than splitting the sentences into single-region pairs \citep{li2024asg}. Such sentence--region associations, including those involving multiple regions, can also be obtained automatically using medical phrase-grounding models \citep{zhang2026medgrounder}.

\citet{dallaserra2023controllable} preserve associations between sentences and multiple anatomical regions, allowing partial reports to be generated jointly from these regions. To align regional inputs with target text, however, they merge overlapping region sets together with their associated sentences. This grouping can expand transitively through shared regions, incorporating additional regions and sentences, thereby yielding coarser region-set--text supervision. At inference, the method conditions generation on externally supplied anatomical subsets.

CURE's anatomy-grounded report generation mode uses predefined anatomical regions as queries, localizes and describes each in turn, and concatenates the outputs. The query set rather than the image determines which units a report contains, with broader enumeration increasing recall but reducing precision \citep{messina2026cure}. ChEX likewise generates descriptions for anatomical queries, but in report generation it first filters the queried regions using image-based classification, similar to RGRG's region selection \citep{muller2025chex}.

RadCarve instead defines generation units directly from the anatomical region sets in existing statement--region correspondences, without single-region projection or expansion through shared regions. Its planner jointly predicts unit existence and region composition from the image, iteratively refining both with evidence from each unit's candidate regions.

\section{Method}
\label{sec:method}

\subsection{Overview}
\label{sec:overview}

As shown in Figure~\ref{fig:overview}, RadCarve consists of a region-set planner and a unit-conditioned text generator. Given a chest X-ray image $I$, the planner predicts nonempty region sets $R_t \subseteq \mathcal R$ from detector and Rad-DINO \citep{perezgarcia2025raddino} representations, where $\mathcal R$ denotes all anatomical region categories. Composition-specialized query banks dedicate queries to different region-composition types, and iterative composition refinement aggregates regional evidence according to current membership predictions to refine both unit existence and region composition. Each $R_t$ defines a generation unit and its regional visual evidence. Conditioned on the set-level representation of $R_t$ and the region-level features of its members, the generator independently produces the unit text $y_t$ (Appendix~\ref{app:representations}).

\subsection{Generation Units and Supervision Construction}
\label{sec:supervision}

Let $\{s_j\}_{j=1}^{M}$ denote the statements in a report, and let $A_j\subseteq\mathcal R$ be the anatomical region set associated with statement $s_j$ in Chest ImaGenome. Each distinct nonempty $A_j$ in a report defines a source unit with region set $B_t$, yielding $N$ source units $\{B_t\}_{t=1}^{N}$.

\paragraph{Anatomical redundancy pruning.}
Chest ImaGenome's automatically generated annotations \citep{wu2021chest} may associate a statement with both a specific region and its broader parent, such as a lower lung zone and the whole lung \citep{zhang2026medgrounder}. Retaining the parent extends the unit's visual evidence beyond the specific region and can introduce statements associated with other subregions into the text target (Table~\ref{tab:unit_construction}). We therefore remove such redundant parent regions before constructing the final text target, retaining the finer-grained regions already present in each annotated set. Pruning thus operates within each set, rather than decomposing it into single-region units or merging it with other sets.

Let $\mathcal H$ be a fixed set of 21 directed anatomical redundancy relations (Appendix~\ref{app:pruning_rules}), where $(c,p)$ links a specific region $c$ to a broader region $p$. When both occur in $B_t$, we retain $c$ and remove $p$, yielding the final region set:
\begin{equation}
R_t^{\mathrm{gt}}
= B_t\setminus\{p\in B_t\mid \exists c\in B_t,\ (c,p)\in\mathcal H\}.
\label{eq:pruning}
\end{equation}
The statements associated with $R_t^{\mathrm{gt}}$ are concatenated to form the final text target $y_t^{\mathrm{gt}}$. The resulting sets $R_t^{\mathrm{gt}}$ supervise both unit existence and region composition in the planner, and $y_t^{\mathrm{gt}}$ provides the generator target (Appendix~\ref{app:supervision_construction}).

\subsection{Region-Set Planning}
\label{sec:planner}

\begin{figure}[t]
  \centering
  \includegraphics[width=\columnwidth]{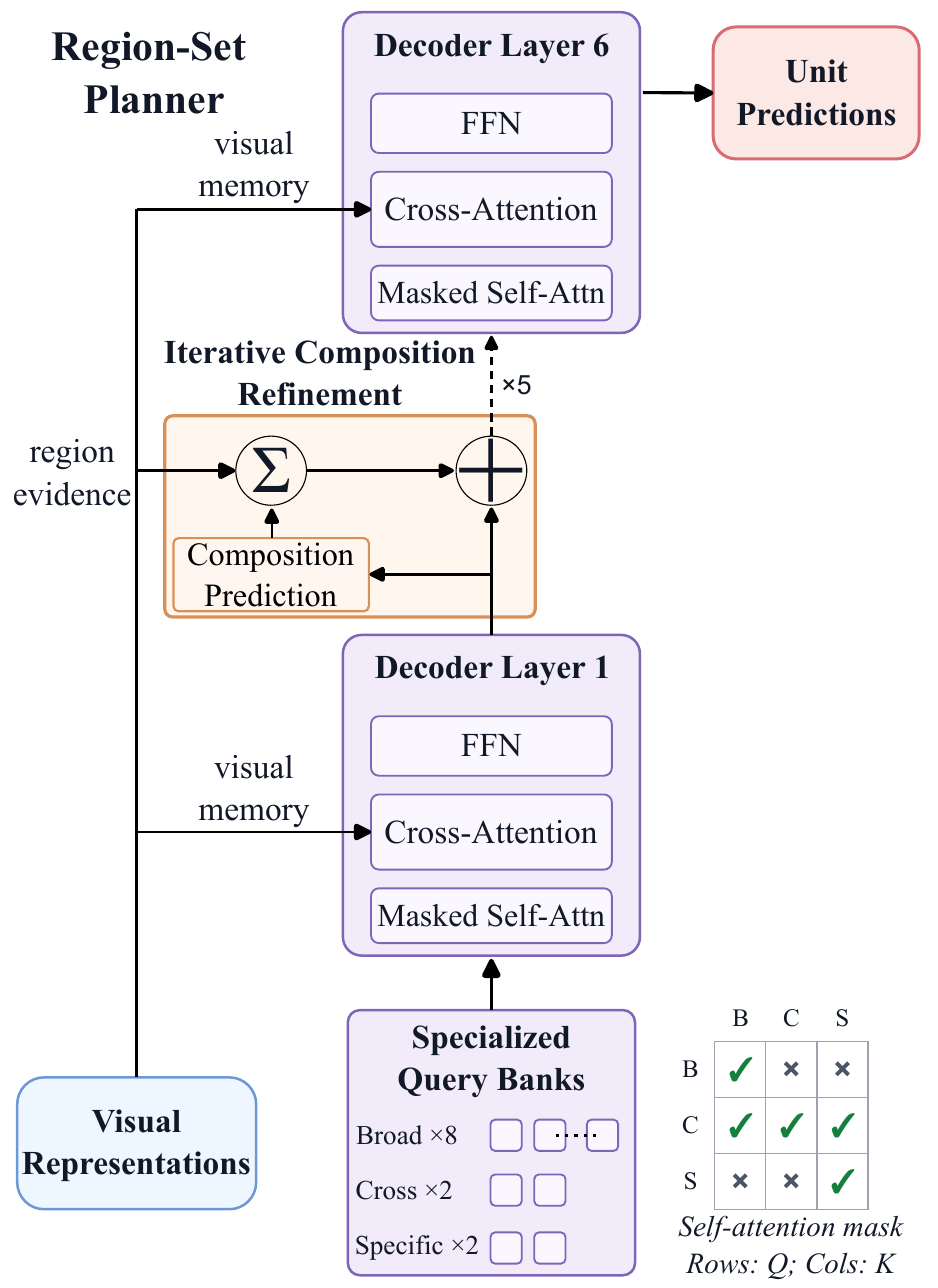}
  \caption{Region-set planner with composition-specialized query banks and iterative composition refinement. The $\times5$ denotes five feedback updates across the six-layer decoder; intermediate layers are omitted. Final predictions specify unit existence and region composition.}
  \label{fig:planner}
\end{figure}

\paragraph{Planner architecture.}
The planner adopts a query-based set prediction framework \citep{carion2020detr}, implemented with a Transformer encoder--decoder \citep{vaswani2017attention}. Each learnable query maintains the set-level representation of a candidate generation unit. Anatomical region representations combine detector ROI features \citep{ren2015fasterrcnn,tanida2023rgrg} with box-aligned local features from the frozen, radiology-pretrained Rad-DINO. A region encoder models inter-region relations, and its outputs, together with global Rad-DINO patch representations, form the decoder's visual memory. As shown in Figure~\ref{fig:planner}, the decoder updates each candidate unit's set-level representation across $L$ layers through cross-attention to this visual memory. The final representation is used to predict both unit existence and region composition.

\paragraph{Composition-specialized query banks.}
Region-set prediction involves compositions with different structural characteristics and learning difficulties. Using training-set diagnostics, we identify a difficult-region subset $\mathcal R_S$ whose members are well detected but poorly recovered by a homogeneous-query planner (Appendix~\ref{app:planner_details}). Most of these regions occur infrequently in training units, limiting positive supervision for their associated compositions. Without composition-type constraints on matching, the same query can be repeatedly matched to structurally different region compositions across training images, potentially hindering stable specialization. We therefore organize queries into composition-specialized banks for three types defined by their relation to $\mathcal R_S$: Broad sets contain no regions in $\mathcal R_S$, Specific sets are nonempty subsets of $\mathcal R_S$, and Cross sets contain regions both inside and outside $\mathcal R_S$.

The three corresponding query banks share the visual encoder and Transformer decoder but have separate prediction heads. One-to-one Hungarian matching is performed between queries and ground-truth sets within each bank \citep{kuhn1955hungarian}. When an image contains no target units of a bank's type, its queries are supervised to predict nonexistence rather than being matched to units of another type. Broad and Specific queries interact only within their respective banks, whereas Cross queries can interact with queries from all banks.

\paragraph{Region-set prediction and iterative composition refinement.}
Region membership decisions are interdependent: whether a region belongs in a set depends not only on its own visual evidence but also on the other candidate members. Beyond modeling inter-region relations in the encoder, we refine each candidate unit's set-level representation by aggregating regional evidence according to its current membership predictions.

Let $h_t^{(\ell)}$ denote the set-level representation of candidate unit $t$ at decoder layer $\ell$, and let $v_r$ be the region encoder output for region $r$. At each layer, $h_t^{(\ell)}$ is used to predict both unit existence and region membership. The membership head outputs sigmoid probabilities $a_{tr}^{(\ell)}$, which weight the regional evidence:
\begin{equation}
c_t^{(\ell)}=\sum_{r\in\mathcal R}a_{tr}^{(\ell)}\,m_{tr}\,v_r,
\label{eq:context}
\end{equation}
where $m_{tr}\in\{0,1\}$ masks undetected regions and restricts aggregation for Specific queries to $\mathcal R_S$. The aggregate then updates the set-level representation through a gated residual:
\begin{equation}
\bar h_t^{(\ell)}=h_t^{(\ell)}
+\sigma(\alpha)\operatorname{LN}\!\left(W_{\mathrm{ref}}c_t^{(\ell)}\right),
\label{eq:refinement}
\end{equation}
where $\alpha$ is a learnable scalar gate parameter. The updated $\bar h_t^{(\ell)}$ enters the next decoder layer to refine both unit existence and region composition.

\paragraph{Training and inference.}
Queries matched to ground-truth units are supervised to predict existence and recover their region composition, while unmatched queries are supervised to predict nonexistence. At inference, candidates predicted to exist by the final decoder layer define the generation units, with their region sets pruned as in Section~\ref{sec:supervision} (Appendix~\ref{app:planner_details}).

\subsection{Unit-Conditioned Text Generation}
\label{sec:generator}

\paragraph{Generation conditions.}
For each predicted region set $R_t$, the generator constructs member-region tokens $z_r$ from the same feature sources as the planner (Appendix~\ref{app:representations}). The final set-level representation $h_t^{(L)}$, from which $R_t$ is predicted, is adapted to the generator's input space as the region-set token $u_t$. The generation condition for this unit is
\begin{equation}
T_t=[u_t,z_{r_{t,1}},\ldots,z_{r_{t,K_t}}],\quad K_t=|R_t|.
\label{eq:tokens}
\end{equation}
We insert trainable gated cross-attention layers \citep{alayrac2022flamingo} into a frozen GPT-2-medium backbone fine-tuned on PubMed abstracts \citep{papanikolaou2020dare} to read $T_t$ and independently generate the text $y_t$ for each unit.

\paragraph{Training and inference.}
With the planner frozen, the generator is trained on both ground-truth and predicted region sets by minimizing autoregressive negative log-likelihood under teacher forcing. At inference, active units are decoded independently, and their texts are assembled into the final report after sentence-level deduplication (Appendix~\ref{app:generator_training}).

\begin{table*}[t]
\centering
\normalsize
\setlength{\tabcolsep}{2pt}
\begin{adjustbox}{max width=\linewidth}
\begin{tabular}{@{}lcccccccccc@{}}
\toprule
Method & \metrichead{Region IoU} & \metrichead{Pair Pass} & \metrichead{B4} & \metrichead{MTR} & \metrichead{R-L} & \metrichead{CXRFES} & \metrichead{CXRFES$_{\geq2}$} & \metrichead{F1$_{\mathrm{mic}\text{-}5}$} & \metrichead{F1$_{\mathrm{micro}\text{-}14}$} & \metrichead{F1$_{\mathrm{ex}\text{-}14}$}\\
\midrule
ChEX & 0.334 & 0.192 & 0.000 & 0.018 & 0.048 & 0.107 & 0.078 & 0.158 & 0.112 & 0.060\\
RGRG & \underline{0.760} & \underline{0.759} & \underline{0.089} & 0.104 & \underline{0.242} & \underline{0.526} & \underline{0.547} & 0.320 & 0.444 & \underline{0.419}\\
CURE & 0.677 & 0.720 & \textbf{0.103} & \underline{0.113} & 0.236 & 0.520 & 0.531 & \underline{0.338} & \underline{0.461} & 0.404\\
RadCarve & \textbf{0.761} & \textbf{0.764} & 0.074 & \textbf{0.119} & \textbf{0.298} & \textbf{0.576} & \textbf{0.581} & \textbf{0.384} & \textbf{0.500} & \textbf{0.448}\\
\bottomrule
\end{tabular}
\end{adjustbox}
\caption{Localization and generation given anatomical region sets. Pair Pass requires all target regions to be localized at $\mathrm{IoU}\geq0.5$; failed pairs receive empty text and remain in scoring. CXRFES$_{\geq2}$ evaluates multi-region pairs. Best/second-best results are bolded/underlined.}
\label{tab:given_sets}
\end{table*}

\begin{table*}[t]
\centering
\normalsize
\setlength{\tabcolsep}{3pt}
\begin{adjustbox}{max width=\linewidth}
\begin{tabular}{@{}lccccccc@{}}
\toprule
Method & \metrichead{Binding IoU} & \metrichead{P@0.5} & \metrichead{R@0.5} & \metrichead{F1@0.5} & Pred/GT & \metrichead{Single IoU} & \metrichead{Multi IoU}\\
\midrule
RGRG & \underline{0.262} & \underline{0.127} & \underline{0.337} & \underline{0.185} & 2.648 & \underline{0.317} & \underline{0.242}\\
ChEX & 0.080 & 0.079 & 0.102 & 0.089 & 1.290 & 0.134 & 0.061\\
RadCarve & \textbf{0.424} & \textbf{0.419} & \textbf{0.489} & \textbf{0.451} & 1.165 & \textbf{0.322} & \textbf{0.462}\\
\bottomrule
\end{tabular}
\end{adjustbox}
\caption{End-to-end binding recovery after text-only matching. IoUs average over targets, with zero for unmatched targets; Single/Multi IoU use single-/multi-region targets. P/R/F1 use a set-IoU threshold of 0.5. Pred/GT is the predicted-to-target unit count ratio. Best/second-best results are bolded/underlined, except for the descriptive Pred/GT ratio.}
\label{tab:binding}
\end{table*}

\begin{table*}[t]
\centering
\normalsize
\setlength{\tabcolsep}{2pt}
\begin{adjustbox}{max width=\linewidth}
\begin{tabular}{@{}lccccccccc@{}}
\toprule
Method & Params (B) & \metrichead{B1} & \metrichead{B4} & \metrichead{MTR} & \metrichead{R-L} & \metrichead{CXRFES} & \metrichead{F1$_{\mathrm{mic}\text{-}5}$} & \metrichead{F1$_{\mathrm{micro}\text{-}14}$} & \metrichead{F1$_{\mathrm{ex}\text{-}14}$}\\
\midrule
ChEX & 1.051 & 0.285 & 0.074 & 0.136 & 0.213 & 0.625 & 0.523 & 0.481 & 0.410\\
RGRG & 0.609 & \underline{0.376} & \underline{0.127} & \underline{0.167} & \underline{0.264} & \underline{0.720} & 0.546 & 0.497 & 0.452\\
MAIRA-2 & 6.880 & 0.198 & 0.034 & 0.103 & 0.177 & 0.626 & \underline{0.551} & \underline{0.526} & \underline{0.483}\\
MAIRA-2 (grounded) & 6.880 & 0.210 & 0.043 & 0.104 & 0.219 & 0.681 & 0.453 & 0.461 & 0.416\\
CURE & 5.010 & 0.095 & 0.036 & 0.156 & 0.145 & 0.705 & 0.544 & 0.458 & 0.386\\
PriorRG & 0.336 & 0.268 & 0.050 & 0.126 & 0.215 & 0.671 & 0.534 & 0.512 & 0.459\\
R2Gen-Mamba & 0.081 & 0.298 & 0.076 & 0.136 & \textbf{0.268} & 0.681 & 0.329 & 0.363 & 0.350\\
RadCarve & 0.761 & \textbf{0.390} & \textbf{0.131} & \textbf{0.169} & \textbf{0.268} & \textbf{0.724} & \textbf{0.566} & \textbf{0.534} & \textbf{0.490}\\
\bottomrule
\end{tabular}
\end{adjustbox}
\caption{Complete-report generation on the Chest ImaGenome processed test set. Params denotes total inference parameters, including frozen weights. Best/second-best results are bolded/underlined.}
\label{tab:indomain}
\end{table*}

\begin{table}[t]
\centering
\normalsize
\setlength{\tabcolsep}{1.5pt}
\begin{adjustbox}{max width=\linewidth}
\begin{tabular}{@{}lcccc@{}}
\toprule
Method & \metrichead{CXRFES} & \metrichead{F1$_{\mathrm{mic}\text{-}5}$} & \metrichead{F1$_{\mathrm{micro}\text{-}14}$} & \metrichead{F1$_{\mathrm{ex}\text{-}14}$}\\
\midrule
ChEX & 0.488 & 0.254 & 0.317 & 0.316\\
RGRG & 0.543 & \underline{0.309} & 0.450 & 0.410\\
PriorRG & \textbf{0.552} & 0.289 & 0.384 & 0.346\\
R2Gen-Mamba & 0.489 & 0.210 & \underline{0.455} & \underline{0.426}\\
RadCarve & \underline{0.546} & \textbf{0.398} & \textbf{0.508} & \textbf{0.445}\\
\bottomrule
\end{tabular}
\end{adjustbox}
\caption{Complete-report generation on the PadChest-GR test set without target-domain training or fine-tuning. Best/second-best results are bolded/underlined.}
\label{tab:external}
\end{table}

\begin{table*}[t]
\centering
\normalsize
\setlength{\tabcolsep}{2pt}
\begin{adjustbox}{max width=\linewidth}
\begin{tabular}{@{}lcccccccc@{}}
\toprule
Method & \metrichead{B1} & \metrichead{B4} & \metrichead{MTR} & \metrichead{R-L} & \metrichead{CXRFES} & \metrichead{F1$_{\mathrm{mic}\text{-}5}$} & \metrichead{F1$_{\mathrm{micro}\text{-}14}$} & \metrichead{F1$_{\mathrm{ex}\text{-}14}$}\\
\midrule
RadCarve & \textbf{0.390} & \textbf{0.131} & \underline{0.169} & \textbf{0.268} & \textbf{0.724} & \textbf{0.566} & \textbf{0.534} & \textbf{0.490}\\
w/o Set Binding & 0.347 & 0.115 & 0.147 & \underline{0.266} & 0.717 & 0.507 & 0.461 & 0.438\\
w/o Pruning & \underline{0.380} & 0.124 & 0.162 & 0.264 & 0.714 & 0.511 & 0.482 & 0.440\\
\quad $\hookrightarrow$ Hybrid$^\dagger$ & 0.364 & 0.116 & \textbf{0.174} & 0.260 & \underline{0.723} & \underline{0.559} & \underline{0.524} & \underline{0.473}\\
w/o Refinement & 0.364 & 0.120 & 0.158 & 0.257 & 0.714 & 0.519 & 0.480 & 0.446\\
w/o Specialized Banks & 0.374 & \underline{0.127} & 0.162 & 0.261 & 0.718 & 0.538 & 0.498 & 0.461\\
\bottomrule
\end{tabular}
\end{adjustbox}
\caption{Complete-report ablations on the Chest ImaGenome processed test set. Best/second-best results are bolded/underlined. $^\dagger$Hybrid replaces only the generator of w/o Pruning with the full RadCarve generator; inference-time pruning remains disabled.}
\label{tab:ablations}
\end{table*}

\section{Results}
\label{sec:results}

\subsection{Experimental Setup}
\label{sec:setup}

\paragraph{Datasets.}
We train on MIMIC-CXR \citep{johnson2019mimiccxr} with Chest ImaGenome annotations \citep{wu2021chest}. All in-domain evaluations draw from the 32,711-image Chest ImaGenome processed test set, with samples included according to each task's annotation requirements (Appendix~\ref{app:dataset_statistics}). For external complete-report evaluation, we use the 915-sample official PadChest-GR test set, which provides finding-level text--box annotations rather than anatomical region-set targets \citep{castro2025padchestgr}.

\paragraph{Baselines.}
We compare with MAIRA-2, a leading large-scale report-generation model \citep{bannur2024maira2}, and CURE, a large-scale model with state-of-the-art visual grounding and grounded report generation performance \citep{messina2026cure}, alongside RGRG \citep{tanida2023rgrg}, ChEX \citep{muller2025chex}, PriorRG \citep{liu2026priorrg}, and R2Gen-Mamba \citep{sun2025r2genmamba}. RGRG and R2Gen-Mamba are trained on the same Chest ImaGenome training split as RadCarve. For ChEX, CURE, MAIRA-2, and PriorRG, we use the officially released models. Complete-report generation uses only the current frontal image. Each comparison uses the same test samples, reference reports, and scoring procedures. Further details are provided in Appendix~\ref{app:baseline_details}.

\paragraph{Metrics.}
We report BLEU-1 (B1) and BLEU-4 (B4) \citep{papineni2002bleu}, METEOR (MTR) \citep{banerjee2005meteor}, ROUGE-L (R-L) \citep{lin2004rouge}, and factual semantic consistency measured by CXRFEScore (CXRFES) \citep{messina2024cxrfe}. Clinical metrics are CheXbert \citep{smit2020chexbert} micro-F1 over five disease labels (F1$_{\mathrm{mic}\text{-}5}$) and all 14 labels (F1$_{\mathrm{micro}\text{-}14}$), and mean per-example F1 over 14 labels (F1$_{\mathrm{ex}\text{-}14}$). We assess uncertainty and pairwise differences using 10,000 paired image-level bootstrap resamples (Appendix~\ref{app:statistical_analysis}).

\subsection{Generation Units and Evidence Binding}
\label{sec:units}

\paragraph{Unit generation given anatomical region sets.}
We compare generation units in terms of text quality and anatomical localization on 156,540 text--region-set pairs from 32,568 test images with ground-truth boxes available for all target members. Methods are given the anatomical regions to describe, but neither target text nor ground-truth boxes. RadCarve bypasses its planner and generates text without the region-set token, while baselines use their native regional interfaces. Because methods may produce plausible regional descriptions without drawing on the corresponding image regions, we apply \emph{complete-pair gating}: text is credited only when all target regions are localized at $\mathrm{IoU}\geq0.5$ (Appendix~\ref{app:given_sets_protocol}).

RadCarve outperforms all baselines, including CURE, on CXRFES and all three clinical F1 metrics (Table~\ref{tab:given_sets}). This advantage holds over RGRG despite their similar localization performance. RadCarve's CXRFES lead persists on multi-region pairs, supporting the joint use of related regions for generation. ChEX's low complete-pair pass rate limits its text scores under the joint evaluation.

\paragraph{End-to-end unit generation and binding recovery.}
Evaluation with supplied region sets does not test whether a model can determine generation units from the image and recover their text--region-set bindings. We therefore evaluate RadCarve, RGRG, and ChEX on 31,271 test images without supplying target region sets. Within each report, we match predicted and target unit texts by CXRFES, then evaluate the corresponding region sets by Jaccard IoU. Text-only matching keeps unit correspondence independent of the region sets being evaluated (Appendix~\ref{app:binding_protocol}).

RadCarve achieves a binding F1@0.5 of 0.451, compared with 0.185 for RGRG and 0.089 for ChEX (Table~\ref{tab:binding}). Despite predicting approximately 2.6 times the target unit count, RGRG achieves lower binding recall than RadCarve, which also attains higher precision with a count closer to the target. Compared with RGRG, RadCarve achieves similar Single IoU but substantially higher Multi IoU, supporting the use of region-set units to recover multi-region bindings. This comparison captures the ability of each method's native generation units to represent and recover the target bindings. RadCarve's overall recovery advantage also holds under region-only matching (Table~\ref{tab:region_only_comparison}). Qualitative examples show the bindings between generated unit texts and their predicted region sets (Appendix~\ref{app:qualitative}).

\begin{figure}[t]
\centering
\includegraphics[width=\columnwidth]{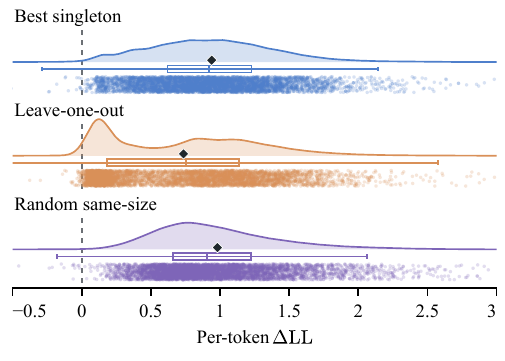}
\caption{Evidence-composition interventions on 86,011 pairs. Dots show per-token log-likelihood gains of Full set over each control for 5,000 randomly sampled pairs per condition. Densities, boxes (quartiles and medians), and diamonds (means) use all pairs. Exact scores and bootstrap confidence intervals are reported in Table~\ref{tab:intervention}.}
\label{fig:intervention}
\end{figure}

\paragraph{Evidence composition intervention.}
To test whether the generator uses region composition, we vary the input region set on 86,011 correctly localized multi-region pairs. We compare Full set with the target region with highest target-text likelihood (Best singleton), one-member deletions averaged over members (Leave-one-out), and equally sized sets of correctly localized nontarget regions from the same image (Random same-size). We measure the per-token log-likelihood gain of Full set over each control for the target text (Appendix~\ref{app:intervention_protocol}).

Full set yields positive gains over both Best singleton and Random same-size, indicating that neither a single region nor region count alone accounts for its advantage (Figure~\ref{fig:intervention}). Removing members (Leave-one-out) also weakens support for the target text. Full set's advantage holds for nearly all evaluated pairs across all three controls. Together, these results show that the generator uses the joint evidence provided by the selected region set.

\subsection{Report Generation Performance}
\label{sec:reports}

\paragraph{In-domain report generation.}
RadCarve achieves the highest CXRFES and all three clinical F1 scores (Table~\ref{tab:indomain}). MAIRA-2 achieves higher clinical scores in its non-grounded mode than in its grounded mode, yet RadCarve outperforms both while providing region-set bindings for its generated units. The gains over the stronger non-grounded mode are significant on all three clinical metrics (Appendix~\ref{app:statistical_analysis}). RadCarve also leads in BLEU-1, BLEU-4, and METEOR, and ties with R2Gen-Mamba for the highest ROUGE-L.

RadCarve achieves this performance with 0.761B inference parameters, compared with 6.880B for MAIRA-2 and 5.010B for CURE. Its report-generation task training uses only MIMIC-CXR data with Chest ImaGenome annotations, whereas MAIRA-2 and CURE train on multiple chest X-ray datasets \citep{bannur2024maira2,messina2026cure}.

\paragraph{External evaluation.}
We evaluate report generation on PadChest-GR without target-domain training or fine-tuning. MAIRA-2 and CURE are excluded because they use PadChest report supervision during training. RadCarve achieves the highest scores on all three clinical metrics (Table~\ref{tab:external}). These results indicate that RadCarve's clinical report-generation performance extends beyond its training dataset.

\paragraph{All-black input.}
On an all-black image, the tested baselines continue to generate clinical descriptions, some asserting positive findings (Appendix~\ref{app:null_input}). ChEX generates 25 units, exceeding its real-image average of 20.6, whereas RadCarve generates only one unit, compared with an average of 5.6 on real images. RadCarve's resulting text receives no CheXbert observation labels. Together with the unit-recovery results on real images, this contraction suggests that the planner adjusts unit generation in response to image content.

\subsection{Ablation Studies}
\label{sec:ablations}

Table~\ref{tab:ablations} summarizes complete-report ablations, and Table~\ref{tab:planner_ablations} reports planner recovery under region-only matching. Configurations and additional diagnostics are provided in Appendix~\ref{app:ablation_diagnostics}.

\begin{table*}[!t]
\begingroup
\centering
\normalsize
\setlength{\tabcolsep}{6pt}
\begin{adjustbox}{max width=\linewidth}
\begin{tabular}{@{}lccccccc@{}}
\toprule
Method & \metrichead{Set IoU} & \metrichead{P@0.5} & \metrichead{R@0.5} & \metrichead{F1@0.5} & Pred/GT & \metrichead{Single IoU} & \metrichead{Multi IoU}\\
\midrule
RadCarve & \textbf{0.567} & \textbf{0.551} & \textbf{0.642} & \textbf{0.593} & 1.165 & \textbf{0.415} & \textbf{0.622}\\
w/o Refinement & \underline{0.553} & 0.533 & \underline{0.630} & \underline{0.577} & 1.182 & \underline{0.387} & \underline{0.613}\\
w/o Specialized Banks & 0.538 & \underline{0.547} & 0.611 & \underline{0.577} & 1.118 & 0.342 & 0.609\\
\bottomrule
\end{tabular}
\end{adjustbox}
\caption{Planner ablations under region-only matching on the same test set as Table~\ref{tab:binding}. Metric definitions and aggregation follow that table. Best/second-best results are bolded/underlined.}
\label{tab:planner_ablations}
\endgroup
\end{table*}

Decomposing predicted sets into unique single-region units and retraining the generator with single-region supervision (w/o Set Binding) reduces all metrics despite preserving anatomical coverage. The clinical decline primarily reflects omissions: R$_{\mathrm{micro}\text{-}14}$ falls from 0.507 to 0.393, while P$_{\mathrm{micro}\text{-}14}$ decreases only slightly from 0.564 to 0.556. The exact duplicate sentence rate also rises from 0.139 to 0.310 (Table~\ref{tab:singleton}). These results support jointly using evidence from related regions within a generation unit to recover more complete clinical content with less redundancy.

We disable anatomical redundancy pruning throughout training and inference and retrain both components (w/o Pruning). Pruning changes the region sets of 58.12\% of training source units and the text targets of 45.78\% (Table~\ref{tab:pruning_impact}). Disabling this step reduces all metrics. To assess the effect of pruned supervision while keeping inference inputs unchanged, Hybrid replaces only the unpruned model's generator with the RadCarve generator trained on pruned supervision. It retains the unpruned model's region sets and region-set tokens, with inference-time pruning still disabled. It raises F1$_{\mathrm{micro}\text{-}14}$ from 0.482 to 0.524, recovering most of the gap to the full model on all three clinical metrics. This recovery indicates that supervision constructed after pruning helps the generator learn more focused region-set--text correspondences.

Removing composition feedback while retaining per-layer predictions and intermediate supervision (w/o Refinement) reduces text and clinical metrics. Under region-only matching, Set IoU falls from 0.567 to 0.553 despite similar predicted unit counts (Table~\ref{tab:planner_ablations}). The full model's advantage persists across all decoder layers (Table~\ref{tab:refinement_layers}), supporting the role of composition feedback in improving set recovery.

Replacing composition-specialized query banks with an equally sized homogeneous query pool (w/o Specialized Banks) reduces all report-generation metrics. Under region-only matching, F1@0.5 drops by 0.064 for Cross and 0.062 for Specific sets, compared with 0.008 for Broad sets (Table~\ref{tab:types}), indicating greater benefits for compositions involving the difficult-region subset.

\FloatBarrier
\section{Conclusion}
\label{sec:conclusion}

We introduce RadCarve, which defines generation units directly from the anatomical region sets in existing statement--region correspondences and predicts their existence and composition from the image. Each region set specifies both a generation unit and the regional evidence used to generate its text. Our evaluations demonstrate improved recovery of multi-region bindings and support the joint use of related regions for more complete clinical content. These benefits extend to complete-report generation, with strong in-domain and external performance. Together, these results suggest that visual evidence is not only a condition for text generation: its composition should also inform the definition of generation units.

\section*{Limitations}

\paragraph{Supervision quality.}
RadCarve derives generation units and text targets from automatically constructed statement--region correspondences in Chest ImaGenome. These annotations may include content not strictly related to the associated regions \citep{messina2026cure}, introducing noise into the resulting supervision. Anatomical redundancy pruning addresses specific parent--child redundancies but does not resolve all region--text mismatches.

\paragraph{Single-image input.}
Our experiments use a single current frontal radiograph, without prior examinations, lateral views, or clinical context. This setting does not provide the longitudinal evidence needed to assess interval changes described in reference reports, and comparative statements generated by the model cannot be verified from the supplied input alone.

\section*{Ethical Considerations}

We use MIMIC-CXR \citep{johnson2019mimiccxr} and Chest ImaGenome \citep{wu2021chest} under their controlled-access data use agreements, and PadChest-GR \citep{castro2025padchestgr} for external evaluation under its applicable terms of use. All authors who accessed controlled-access data completed the required credentialing and data use agreements. RadCarve is intended for research on assisted report generation; clinical use would require further validation and radiologist oversight.

\bibliography{custom}

\clearpage
\appendix
\section{Generation Units and Supervision Construction}
\label{app:supervision_details}

\subsection{Statement--Region Correspondences and Supervision Construction}
\label{app:supervision_construction}

We obtain statement--region correspondences from the phrases attached to anatomical regions in the Chest ImaGenome silver scene graphs \citep{wu2021chest}. After text cleaning, phrases with the same normalized text are merged into a statement $s_j$, whose associated set $A_j$ contains all regions linked to that text. Matching ignores case, repeated whitespace, and trailing periods; the first encountered text form is retained. Statements are ordered by their earliest annotation phrase IDs. Here, a statement follows the annotation phrase granularity and may contain more than one grammatical sentence.

Each distinct nonempty $A_j$ defines a source unit with region set $B_t$. We prune each $B_t$ independently to obtain $R_t^{\mathrm{gt}}$ (Equation~\ref{eq:pruning}), retaining separate training units even when their pruned sets coincide.

Following prior work \citep{tanida2023rgrg,messina2026cure}, the statements associated with $R_t^{\mathrm{gt}}$ are concatenated in annotation order to form the text target $y_t^{\mathrm{gt}}$. Thus, generation units are defined directly from the region sets in existing statement--region correspondences, whereas the statements associated with those sets determine the text targets.

Table~\ref{tab:unit_construction} compares three training-unit constructions in a Chest ImaGenome example. The automatically derived statement--region associations may include content not strictly related to the associated regions \citep{messina2026cure}. RGRG projects multi-region correspondences onto individual regions \citep{tanida2023rgrg}, assigning ``The lungs are clear'' to each single-lung unit. The grouping of \citet{dallaserra2023controllable} merges overlapping region sets through shared regions, producing one unit containing six regions and all four statements. RadCarve keeps four separate region-set units, removing the mediastinum region only when a more specific region occurs in the same set. These units accommodate both cross-region descriptions and localized findings.

\begin{table*}[p]
\begingroup
\centering
\begin{tabularx}{\linewidth}{@{}p{0.12\linewidth}>{\raggedright\arraybackslash}p{0.40\linewidth}>{\raggedright\arraybackslash}X@{}}
\toprule
Method & Regional input & Training text\\
\midrule
RGRG & \{right lung\} & The lungs are clear. No signs of pneumonia or CHF.\\
\addlinespace
 & \{left lung\} & The lungs are clear. No signs of pneumonia or CHF.\\
\addlinespace
 & \{cardiac silhouette\} & No signs of pneumonia or CHF. Cardiomediastinal silhouette appears normal.\\
\addlinespace
 & \{mediastinum\} & Cardiomediastinal silhouette appears normal. The aorta is slightly unfolded.\\
\addlinespace
 & \{upper mediastinum\} & Cardiomediastinal silhouette appears normal.\\
\addlinespace
 & \{aortic arch\} & The aorta is slightly unfolded.\\
\midrule
Dalla Serra & \{right lung, left lung, cardiac silhouette, mediastinum, upper mediastinum, aortic arch\} & The lungs are clear. No signs of pneumonia or CHF. Cardiomediastinal silhouette appears normal. The aorta is slightly unfolded.\\
\midrule
RadCarve w/o pruning & \{mediastinum, upper mediastinum, cardiac silhouette\} & No signs of pneumonia or CHF. Cardiomediastinal silhouette appears normal. The aorta is slightly unfolded.\\
\addlinespace
 & \{mediastinum, aortic arch\} & Cardiomediastinal silhouette appears normal. The aorta is slightly unfolded.\\
\midrule
RadCarve & \{right lung, left lung\} & The lungs are clear. No signs of pneumonia or CHF.\\
\addlinespace
 & \{right lung, left lung, cardiac silhouette\} & The lungs are clear. No signs of pneumonia or CHF. Cardiomediastinal silhouette appears normal.\\
\addlinespace
 & \{upper mediastinum, cardiac silhouette\} & No signs of pneumonia or CHF. Cardiomediastinal silhouette appears normal.\\
\addlinespace
 & \{aortic arch\} & The aorta is slightly unfolded.\\
\bottomrule
\end{tabularx}
\caption{Comparison of training-unit constructions in a Chest ImaGenome example. RadCarve w/o pruning shows only the two units affected by pruning.}
\label{tab:unit_construction}
\endgroup
\end{table*}

\begin{table*}[p]
\begingroup
\centering
\begin{adjustbox}{max width=\linewidth}
\begin{tabular}{@{}llcrrr@{}}
\toprule
Child & Parent removed & Type & $P(p\mid c)$ & Coverage & Target changed\\
\midrule
right upper lung zone & right lung & D & 97.83 & 99.70 & 52.94\\
right mid lung zone & right lung & D & 99.80 & 99.85 & 60.35\\
right lower lung zone & right lung & D & 99.04 & 99.83 & 40.43\\
right hilar structures & right lung & E & 77.45 & 99.69 & 23.28\\
right apical zone & right lung & D & 99.50 & 99.71 & 36.88\\
right costophrenic angle & right lung & D & 99.96 & 44.60 & 25.67\\
left upper lung zone & left lung & D & 99.85 & 99.75 & 44.58\\
left mid lung zone & left lung & D & 99.81 & 99.89 & 60.33\\
left lower lung zone & left lung & D & 98.89 & 99.84 & 45.09\\
left hilar structures & left lung & E & 78.28 & 99.85 & 25.30\\
left apical zone & left lung & D & 99.63 & 99.74 & 33.07\\
left costophrenic angle & left lung & D & 99.96 & 50.19 & 27.39\\
right hemidiaphragm & abdomen & E & 62.77 & 86.32 & 6.75\\
left hemidiaphragm & abdomen & E & 50.67 & 92.14 & 9.73\\
aortic arch & mediastinum & A & 99.71 & 99.66 & 46.85\\
SVC & mediastinum & A & 99.27 & 99.78 & 60.87\\
cavoatrial junction & mediastinum & A & 99.60 & 99.69 & 60.03\\
right atrium & mediastinum & A & 98.87 & 99.61 & 41.66\\
upper mediastinum & mediastinum & D & 95.24 & 99.85 & 33.67\\
right atrium & cardiac silhouette & D & 93.88 & 99.22 & 60.18\\
carina & trachea & D & 99.48 & 99.36 & 19.16\\
\bottomrule
\end{tabular}
\end{adjustbox}
\caption{Fixed redundancy rules and training-set diagnostics (\%). D/A: direct/ancestor relations in AOR's anatomical hierarchy; E: dataset-specific overlap extensions. $P(p\mid c)$: parent co-occurrence among source units containing the child. Coverage: fraction of the child's ground-truth box area covered by the parent box. Target changed: fraction of co-occurring source units whose text target changes when only that rule's parent is removed.}
\label{tab:pruning_rules}
\endgroup
\par\smallskip
\begingroup
\centering
\begin{adjustbox}{max width=\linewidth}
\begin{tabular}{@{}lrrrrrr@{}}
\toprule
Split & \shortstack[r]{Source\\units} & \shortstack[r]{Region set\\changed (\%)} & \shortstack[r]{Text target\\changed (\%)} & Mean size & $|R|\geq2$ (\%) & $|R|\geq3$ (\%)\\
\midrule
Train & 758,717 & 58.12 & 45.78 & $3.015\to2.125$ & $85.15\to71.31$ & $54.36\to23.66$\\
Validation & 75,194 & 57.80 & 45.74 & $3.062\to2.177$ & $86.34\to73.69$ & $57.56\to25.62$\\
Test & 151,659 & 57.89 & 45.44 & $3.058\to2.175$ & $86.22\to73.42$ & $57.48\to25.63$\\
\bottomrule
\end{tabular}
\end{adjustbox}
\caption{Effects of anatomical redundancy pruning on supervision and region-set cardinality. Arrows indicate before $\to$ after pruning; percentages are over source units. Test statistics use the 31,271-image fully annotated subset.}
\label{tab:pruning_impact}
\endgroup
\end{table*}

\subsection{Anatomical Redundancy Pruning}
\label{app:pruning_rules}

We define a fixed set $\mathcal H$ of 21 redundancy relations over the 29 anatomical regions in Chest ImaGenome (Table~\ref{tab:pruning_rules}). Seventeen are supported by the anatomical object hierarchy presented in AOR \citep[Figure~A3]{li2025aor}, comprising 13 direct relations and four ancestor relations. We additionally include four dataset-specific overlap relations linking the hilar structures to the ipsilateral lung and the hemidiaphragms to the abdomen, based on their annotation co-occurrence and spatial overlap in the training set. All rules are applied within individual sets as in Equation~\ref{eq:pruning}: a parent is removed only when its corresponding child is also present. Thus, a set containing only the right lung is retained, as is a bilateral lung set.

The hilar regions are almost entirely covered by their ipsilateral lung boxes (99.69\%--99.85\% coverage), motivating retention of the more localized hilar label when both occur in a unit. For the hemidiaphragm--abdomen relations, box coverage is 86.32\%--92.14\%; moreover, removing the abdomen changes the text target in only 6.75\%--9.73\% of source units containing both labels. Thus, the abdomen usually contributes no additional statements to the text target in these units. Together, these spatial and text-target observations support retaining the hemidiaphragm rather than the broader abdomen label.

Removing redundant parent regions can reduce noise in text supervision by excluding statements introduced solely through those regions. In Table~\ref{tab:unit_construction}, removing mediastinum from \{mediastinum, aortic arch\} excludes the cardiomediastinal statement from the text target, leaving only the aortic description.

\subsection{Unit and Supervision Statistics}
\label{app:unit_statistics}

Table~\ref{tab:pruning_impact} summarizes the effects of pruning, counting source units as defined in Appendix~\ref{app:supervision_construction} and retaining one row per source unit as in training. Text comparisons ignore case and leading, trailing, or repeated whitespace. Pruning changes 58.12\% of training region sets but only 45.78\% of text targets, indicating that removing a parent does not always alter the text target. Mean set cardinality decreases from 3.015 to 2.125, with similar reductions across splits.

Of the 758,717 training source units, the number containing at least two regions decreases from 646,059 (85.15\%) to 541,061 (71.31\%). The corresponding proportions decrease from 86.34\% to 73.69\% on validation and from 86.22\% to 73.42\% on test. Multi-region units therefore remain the majority after redundancy removal.

\begin{table*}[t]
\centering
\normalsize
\setlength{\tabcolsep}{4pt}
\begin{tabularx}{\linewidth}{@{}lX@{}}
\toprule
Component & Configuration\\
\midrule
Anatomical detector & ResNet-50 Faster R-CNN; 29 regions; $512\times512$ input; 1024-dimensional ROI features\\
Rad-DINO & 12-layer ViT; width 768; patch size 14; $448\times448$ input; CLS and $32\times32$ patch tokens\\
Planner region encoder & 29 region tokens; width 256; 2 layers; 4 attention heads\\
Planner query decoder & Width 256; 6 layers; 4 attention heads; 12 queries (Broad/Cross/Specific: 8/2/2); iterative refinement enabled\\
Language model & GPT-2-medium; 24 layers; width 1024; 16 attention heads\\
Generator visual interface & 1024-dimensional region and region-set tokens; trainable gated visual cross-attention in all 24 language-model layers\\
\bottomrule
\end{tabularx}
\caption{Architecture configuration of the full RadCarve model.}
\label{tab:model_configuration}
\end{table*}

\section{Implementation and Training Details}
\label{app:method_details}

\subsection{Model Configuration and Feature Representations}
\label{app:representations}
Table~\ref{tab:model_configuration} summarizes the full model. We use ResNet-50 Faster R-CNN for anatomical region detection \citep{ren2015fasterrcnn,tanida2023rgrg}, Rad-DINO for complementary visual features \citep{perezgarcia2025raddino}, and a PubMed-adapted GPT-2-medium model for text generation \citep{papanikolaou2020dare}.\footnote{Model checkpoints: Rad-DINO, \url{https://huggingface.co/microsoft/rad-dino}; PubMed-adapted GPT-2-medium, \url{https://huggingface.co/healx/gpt-2-pubmed-medium}.}

Both the planner and the generator use detector region features, obtained by spatially averaging the ROI feature map of the highest-scoring proposal for each anatomical class and projecting it to 1024 dimensions.

\paragraph{Planner visual representations.}
The planner combines detector region features with gated local Rad-DINO features to construct its region-encoder inputs. The local features combine information from within the detected box and from its surrounding context within a $1.5\times$ expanded box. Rad-DINO CLS features provide image-conditioned initialization of the learnable queries, while patch tokens are concatenated with the encoded region tokens to form the decoder's visual memory. Each query maintains a candidate unit's set-level representation, which is updated across decoder layers using this visual memory and the composition feedback described in Section~\ref{sec:planner}.

\paragraph{Generator visual conditioning.}
For each selected region, the generator extracts local Rad-DINO features only within the detected box using RoIAlign with a $4\times4$ output. These features are fused with detector region features through gating to construct the region token $z_r$.

The planner's final set-level representation $h_t^{(L)}$, from which $R_t$ is predicted, is adapted to the generator's input space through projection, layer normalization, and gating to obtain the region-set token $u_t$. This token is prepended to the member-region tokens to form $T_t$ in Equation~\ref{eq:tokens}.

\subsection{Region-Set Planner Details}
\label{app:planner_details}

\begin{table*}[!t]
\centering
\normalsize
\setlength{\tabcolsep}{8pt}
\begin{adjustbox}{max width=\linewidth}
\begin{tabular}{@{}lrrrc@{}}
\toprule
Anatomical region & GT frequency (\%) & Regional F1 & Detector recall (\%) & Selected\\
\midrule
Right lung & 21.40 & 0.751 & 100.0 & ---\\
Right upper lung zone & 1.68 & 0.444 & 99.8 & $\checkmark$\\
Right mid lung zone & 2.93 & 0.507 & 94.2 & $\checkmark$\\
Right lower lung zone & 8.88 & 0.508 & 96.7 & ---\\
Right hilar structures & 16.40 & 0.596 & 95.4 & ---\\
Right apical zone & 2.51 & 0.517 & 97.2 & $\checkmark$\\
Right costophrenic angle & 17.06 & 0.489 & 43.8 & ---\\
Right hemidiaphragm & 2.45 & 0.491 & 91.8 & $\checkmark$\\
Left lung & 21.13 & 0.750 & 100.0 & ---\\
Left upper lung zone & 1.33 & 0.373 & 99.8 & $\checkmark$\\
Left mid lung zone & 2.98 & 0.494 & 94.4 & $\checkmark$\\
Left lower lung zone & 9.98 & 0.542 & 97.1 & ---\\
Left hilar structures & 16.07 & 0.588 & 92.7 & ---\\
Left apical zone & 2.32 & 0.492 & 97.1 & $\checkmark$\\
Left costophrenic angle & 17.33 & 0.417 & 33.6 & ---\\
Left hemidiaphragm & 1.91 & 0.460 & 91.3 & $\checkmark$\\
Trachea & 1.97 & 0.364 & 94.4 & $\checkmark$\\
Spine & 3.76 & 0.567 & 99.8 & ---\\
Right clavicle & 3.06 & 0.476 & 94.2 & $\checkmark$\\
Left clavicle & 3.04 & 0.479 & 94.6 & $\checkmark$\\
Aortic arch & 2.25 & 0.341 & 68.1 & ---\\
Mediastinum & 11.64 & 0.630 & 100.0 & ---\\
Upper mediastinum & 9.30 & 0.469 & 99.7 & ---\\
SVC & 1.86 & 0.433 & 84.4 & $\checkmark$\\
Cardiac silhouette & 23.28 & 0.640 & 99.9 & ---\\
Cavoatrial junction & 0.50 & 0.001 & 20.1 & ---\\
Right atrium & 0.82 & 0.200 & 68.4 & ---\\
Carina & 1.35 & 0.001 & 0.0 & ---\\
Abdomen & 3.26 & 0.524 & 100.0 & ---\\
\bottomrule
\end{tabular}
\end{adjustbox}
\caption{Per-region training diagnostics of the homogeneous-query planner. GT frequency is measured over source units; detector recall over reports annotated with each region. Regional F1 measures membership recovery after region-only matching. Checkmarks identify the 12 regions in $\mathcal R_S$.}
\label{tab:region_diagnostics}
\end{table*}

\paragraph{Difficult regions and bank configuration.}
Table~\ref{tab:region_diagnostics} reports per-region recovery of a homogeneous-query planner on 166,504 training reports containing 758,717 pruned source units. A multi-region unit contributes to the frequency of each member. Detector recall counts detection of an annotated region, independently of planner recovery. Low detector recall directly limits recovery for some regions, since undetected regions are excluded from predicted sets. We therefore focus on regions with low regional F1 despite high detector recall. Most of these regions occur infrequently in training target units, limiting positive supervision for their associated compositions. Without composition-type constraints on matching, the same query can be repeatedly matched to structurally different region compositions across training images, potentially hindering stable specialization. The selected regions have GT frequency below 4\%, regional F1 below 0.52, and detector recall of at least 80\%. We define $\mathcal R_S$ from these regions and use it to distinguish Broad, Cross, and Specific composition types. The subset comprises 12 regions: the bilateral upper lung zones, mid lung zones, apical zones, hemidiaphragms, and clavicles, together with the trachea and SVC.

The Broad, Cross, and Specific banks contain eight, two, and two queries, respectively. Specific queries attend only to region tokens in $\mathcal R_S$ while retaining access to the full Rad-DINO patch memory. All three membership heads output probabilities over all 29 anatomical regions. Projection, normalization, and gating parameters in iterative composition refinement are shared across feedback updates.

\paragraph{Matching and optimization.}
At every decoder layer, Hungarian matching minimizes negative soft Jaccard similarity within each bank over detected regions, excluding ground-truth sets with no detected members. Matched Broad queries use weighted binary cross-entropy (BCE) for membership prediction, while Cross and Specific queries use a combination of soft-F1 and BCE losses. Unmatched queries receive all-zero membership targets. Existence is supervised with BCE, using positive targets for matched queries and negative targets for the others. Both membership and existence supervision are applied at all six decoder layers.

The planner is trained using AdamW with an initial learning rate of $2.5\times10^{-5}$ and an effective batch size of 128 images across two GPUs. The Rad-DINO backbone remains frozen during planner training. The planner checkpoint is selected based on validation region-set F1@0.5.

\paragraph{Inference.}
We use the final decoder layer's existence and membership predictions. For candidates predicted to exist, we remove undetected regions and apply the anatomical redundancy pruning in Section~\ref{sec:supervision}. Each remaining nonempty region set defines a generation unit.

\subsection{Generator Training and Inference}
\label{app:generator_training}

\paragraph{Training conditions.}
During generator training, the detector, planner (including its Rad-DINO adapters), Rad-DINO backbone, and GPT-2 backbone and language-model head are frozen. The generator's region-token construction, region-set-token mapping, and gated visual cross-attention layers remain trainable.

Ground-truth-set batches use the supervision pairs constructed in Section~\ref{sec:supervision} and condition generation on region tokens only. Predicted-set batches use planner outputs after filtering undetected regions and applying redundancy pruning, including their region-set tokens. Within each composition type, one-to-one matching maximizes region-set Jaccard IoU, and matched pairs with overlapping region sets contribute to the text loss. Each predicted set supplies the visual input, while its matched ground-truth unit supplies the text target. Units with empty visual inputs are skipped.

\paragraph{Curriculum and optimization.}
The proportion of predicted-set batches starts at $1/9$ in the first epoch and increases by $1/9$ per epoch, reaching one in the ninth epoch under the full schedule. The selected checkpoint falls in the fourth epoch, when the proportion is $4/9$. Validation always uses predicted sets, and checkpoint selection is based on validation F1$_{\mathrm{ex}\text{-}14}$. Table~\ref{tab:generator_training_hyperparameters} summarizes the training hyperparameters.

\begin{table}[!htbp]
\centering
\normalsize
\setlength{\tabcolsep}{6pt}
\begin{tabular}{@{}lr@{}}
\toprule
Hyperparameter & Setting\\
\midrule
Optimizer & AdamW\\
Learning rate & $5\times10^{-5}$\\
Weight decay & $0.01$\\
Batch size & $8$\\
Gradient accumulation & $9$ batches\\
Maximum target length & $192$ tokens\\
\bottomrule
\end{tabular}
\caption{Generator-training hyperparameters.}
\label{tab:generator_training_hyperparameters}
\end{table}

\paragraph{Decoding and report assembly.} 
At inference, we decode each predicted unit independently with four-beam search and a maximum length of 300 tokens. We concatenate nonempty unit texts in query order and split the result into sentences. We then remove duplicate sentences in two stages: first, we retain only the first occurrence of each normalized exact duplicate; second, for sentence pairs with BERTScore F1 greater than $0.9$ \citep{zhang2020bertscore}, computed using \texttt{distilbert-base-uncased}, we retain the longer sentence. The remaining sentences form the final report.

\section{Evaluation Details and Supplementary Results}
\label{app:evaluation_protocols}

\subsection{Dataset Statistics}
\label{app:dataset_statistics}
After preprocessing, the MIMIC-CXR data with Chest ImaGenome annotations contain 166,504 training images and 15,658 validation images. All in-domain evaluations draw from the 32,711-image Chest ImaGenome processed test set. Complete report generation uses all 32,711 images. Generation given region sets uses 156,540 eligible pairs from 32,568 images, while end-to-end unit-binding evaluation uses the 31,271 images with complete anatomical region annotations. For complete-report evaluation on Chest ImaGenome, the reference text is the Findings section of the corresponding MIMIC-CXR report. The task-specific inclusion criteria are detailed below. External evaluation uses the 915 samples in the official PadChest-GR test set, without target-domain training or fine-tuning. PadChest-GR provides finding-level text--box annotations rather than anatomical region-set targets, so we use it for complete-report evaluation.

\subsection{Baseline Sources and Training Data}
\label{app:baseline_details}
RGRG, ChEX, CURE, MAIRA-2, and PriorRG use their released models \citep{tanida2023rgrg,muller2025chex,messina2026cure,bannur2024maira2,liu2026priorrg}. The CURE result in the complete-report comparison is from CURE-AGRG-29+GRG. We evaluate MAIRA-2 in both its official grounded and non-grounded report-generation modes using the same released checkpoint. We do not include \citet{dallaserra2023controllable} in the quantitative comparisons because we could not locate publicly available code or pretrained weights for their method.

RadCarve, RGRG, and R2Gen-Mamba use the same Chest ImaGenome training split. RadCarve and RGRG use Chest ImaGenome anatomical annotations, whereas R2Gen-Mamba is trained on complete reports. The remaining baselines use their respective published training data, with ChEX and CURE also using Chest ImaGenome anatomical region--text annotations \citep{muller2025chex,messina2026cure}. In particular, the released ChEX, CURE, MAIRA-2, and PriorRG models use training images from the official MIMIC-CXR training split \citep{johnson2019mimiccxrjpg}, as specified in their respective training protocols \citep{muller2025chex,messina2026cure,bannur2024maira2,liu2026priorrg}. This training split overlaps with the Chest ImaGenome processed test set and may favor these baselines in both given-set and complete-report comparisons, where applicable. Because only 556 images (1.7\%) in the Chest ImaGenome processed test set also belong to the official MIMIC-CXR test split, we use the full processed test set, subject to task-specific inclusion criteria, to retain broader coverage of test images and anatomical compositions.

MAIRA-2 and CURE are excluded from the PadChest-GR comparison because they use PadChest report supervision during training \citep{bannur2024maira2,messina2026cure}. In complete-report comparisons, all methods receive only the current frontal image, without prior images, prior reports, or clinical context. Each comparison uses the same test samples, reference reports, and scoring procedures.

\begin{table}[!htbp]
\centering
\normalsize
\setlength{\tabcolsep}{6pt}
\begin{tabular}{@{}lc@{}}
\toprule
Method & Latency (s/report) $\downarrow$\\
\midrule
ChEX & 1.33\\
RGRG & 1.98\\
MAIRA-2 & 4.21\\
MAIRA-2 (grounded) & 5.12\\
CURE & 65.42\\
PriorRG & 1.41\\
R2Gen-Mamba & 13.44\\
RadCarve & 3.58\\
\bottomrule
\end{tabular}
\caption{Mean complete-report inference latency on an A100 with image batch size one.}
\label{tab:inference_latency}
\end{table}

\paragraph{Inference latency.}
We measure mean report-generation latency on the same 200 test images using an A100 GPU with image batch size one and each method's report-generation configuration. We first run 50 of these images as warm-up, excluding warm-up, model loading, and image reading from the measured time (Table~\ref{tab:inference_latency}).

CURE is evaluated in its AGRG-29+GRG configuration, which generates a separate localization and description for each of 29 anatomical queries and an additional grounded report. The 29 AGRG queries are processed as a batch before the GRG call, and the latency includes both stages.

\subsection{Unit Generation Given Anatomical Region Sets}
\label{app:given_sets_protocol}
We construct text--region-set pairs as in Section~\ref{sec:supervision} across the full processed test set. Pairs require nonempty target text and a nonempty region set, with valid ground-truth boxes for every member. Images need not have annotations for all 29 anatomical regions. Duplicate pairs with identical normalized text and pruned region sets are removed within each image. This yields 156,540 pairs from 32,568 images, including 114,403 multi-region pairs and 339,121 region queries.

All methods are given the anatomical regions to describe, but neither target text nor ground-truth boxes. RadCarve bypasses the planner and conditions generation on the detected target regions' tokens without a region-set token, as in ground-truth-set training batches. RGRG, CURE, and ChEX use their native interfaces to localize and describe each region \citep{tanida2023rgrg,messina2026cure,muller2025chex}, then concatenate the text. Region IoU averages box IoU over all queried regions, assigning zero to undetected regions. Pair Pass is the fraction of pairs for which all target regions are localized at $\mathrm{IoU}\geq0.5$. Text evaluation applies complete-pair gating: failed pairs receive empty candidate text and remain in evaluation. CXRFES$_{\geq2}$ is computed on multi-region pairs. For clinical scoring, CheXbert is applied separately to each pair's complete candidate and reference texts.

\begin{table*}[t]
\centering
\normalsize
\setlength{\tabcolsep}{6pt}
\begin{adjustbox}{max width=\textwidth}
\begin{tabular}{@{}lcccccc@{}}
\toprule
& \multicolumn{3}{c}{Real images} & \multicolumn{3}{c}{All-black input}\\
\cmidrule(lr){2-4}\cmidrule(lr){5-7}
Method & Text response & Complete boxes & Pair Pass & Text response & Complete boxes & Pair Pass\\
\midrule
ChEX & 100.00 & 71.69 & 19.74 & 100.00 & 80.39 & 17.98\\
RGRG & 91.94 & 79.13 & 75.85 & 55.86 & 24.74 & 0.55\\
CURE & 100.00 & 100.00 & 71.10 & 100.00 & 100.00 & 29.11\\
RadCarve & 91.94 & 79.76 & 76.94 & 52.12 & 22.13 & 2.77\\
\bottomrule
\end{tabular}
\end{adjustbox}
\caption{Given-set diagnostic on real and all-black inputs. All values are percentages of target pairs. Text response: nonempty generated text before gating; Complete boxes: localization outputs for every target member; Pair Pass: all target members localized at $\mathrm{IoU}\geq0.5$.}
\label{tab:given_sets_black}
\end{table*}

\paragraph{All-black-input diagnostic.}
We evaluate the same 2,381 target region sets from 500 test images under real and all-black inputs, covering 5,203 member-region queries (Table~\ref{tab:given_sets_black}). Target region sets and reference boxes are unchanged; reference boxes are used only for evaluation. All methods follow the given-set protocol above, with RadCarve bypassing its planner. Text response is measured before gating, and all rates use the full 2,381 pairs as the denominator.

All methods still produce text on all-black inputs. Requiring correct localization excludes many pairs with complete box outputs: for RGRG and RadCarve, the black-input rates fall from 24.74\% and 22.13\% to 0.55\% and 2.77\%, respectively. However, CURE still passes on 29.11\% of pairs, and ChEX's 17.98\% pass rate remains close to its real-image rate of 19.74\%. Complete-pair gating therefore imposes a stricter evaluation condition than merely requiring localization outputs, but passing it does not establish visual support for the generated content.

\subsection{End-to-End Binding Recovery}
\label{app:binding_protocol}
RadCarve, RGRG, and ChEX generate units without target region sets on 31,271 test images containing 151,307 target units. For this evaluation, duplicate target pairs with identical normalized text and pruned region sets are removed within each report. ChEX generates units using its official anatomical-region queries with image-dependent filtering. Each retained query defines a singleton generation unit before text deduplication. We apply the CXRFEScore fact-level scoring function \citep{messina2024cxrfe} to each predicted--target unit pair to obtain the affinity for text-based matching.

For a predicted unit text $p$ and a target unit text $g$, let $F_p$ and $F_g$ denote the sets of radiology facts extracted from these texts using the T5 Fact Extractor. The Chest X-ray Fact Encoder (CXRFE) maps each fact to a 128-dimensional embedding.\footnote{Model checkpoints: T5 Fact Extractor, \texttt{pamessina/T5FactExtractor}; CXRFE, \texttt{pamessina/CXRFE}.} Let $e_i^p$ and $e_j^g$ denote the embeddings of the $i$-th fact in $F_p$ and the $j$-th fact in $F_g$, respectively. We compute the full matrix of pairwise cosine similarities and aggregate it as
\begin{equation}
\begin{aligned}
S(p,g)=\frac{1}{2}\Bigg[
&\frac{1}{|F_p|}\sum_{i=1}^{|F_p|}\max_{1\leq j\leq |F_g|}\cos(e_i^p,e_j^g)\\
+{}&\frac{1}{|F_g|}\sum_{j=1}^{|F_g|}\max_{1\leq i\leq |F_p|}\cos(e_i^p,e_j^g)
\Bigg].
\end{aligned}
\label{eq:unit_matching_affinity}
\end{equation}
Thus, $S(p,g)$ averages the best-match cosine similarity in both fact-matching directions. Within each report, threshold-free Hungarian matching maximizes the sum of $S(p,g)$ over matched unit pairs. All unit texts used for matching yielded usable fact representations, so no unit text was excluded for lacking a usable fact representation.

Region-set Jaccard IoU is computed only after text matching. Binding IoU is the sum of matched-set IoUs divided by the total target unit count, with unmatched targets contributing zero. Single IoU and Multi IoU use the same matching and average over single- and multi-region targets, respectively. Precision, recall, and F1 use a matched-set IoU threshold of 0.5, pooling true positives and predicted and target unit counts across images. Pred/GT is the ratio of predicted to target unit counts.

\paragraph{Region-only matching.}
For text-independent region-set recovery, predicted and target region sets are matched within each report using Hungarian matching to maximize total Jaccard IoU. IoUs use the same target-normalized aggregation as the binding evaluation. Tables~\ref{tab:region_only_comparison} and~\ref{tab:planner_ablations} report method comparisons and planner ablations, respectively, under this protocol.

\begin{figure*}[!t]
\centering
\includegraphics[width=\textwidth]{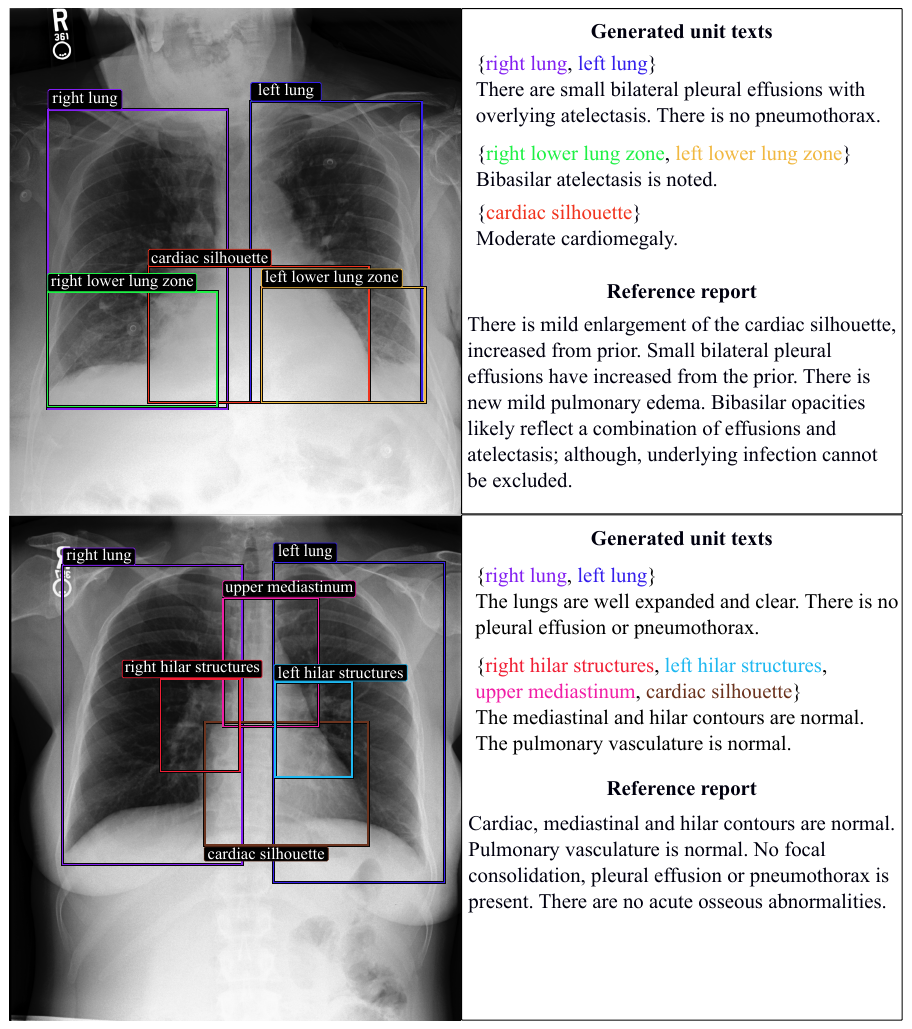}
\caption{Qualitative examples from the test set. Colored boxes mark detected anatomical regions, with matching colors in the selected predicted region sets. Generated unit texts are shown alongside the complete reference report.}
\label{fig:qualitative}
\end{figure*}

\begin{table*}[!t]
\centering
\normalsize
\setlength{\tabcolsep}{5pt}
\begin{adjustbox}{max width=\linewidth}
\begin{tabular}{@{}lcccc@{}}
\toprule
Condition & NLL\,$\downarrow$ & \metrichead{$\Delta$LL} & 95\% CI & \metrichead{$\Delta$LL $>0$}\\
\midrule
Full set & \textbf{0.899} & --- & --- & ---\\
Best singleton & 1.838 & 0.940 & [0.936, 0.943] & 0.996\\
Leave-one-out & \underline{1.635} & 0.736 & [0.732, 0.739] & 0.992\\
Random same-size & 1.880 & 0.981 & [0.978, 0.985] & 0.999\\
\bottomrule
\end{tabular}
\end{adjustbox}
\caption{Evidence-composition interventions (Figure~\ref{fig:intervention}). NLL is mean per-token negative log-likelihood; $\Delta$LL is Full set's mean per-token log-likelihood gain over each control. We report 95\% CIs and the fraction of pairs with positive gains. Best/second-best NLL results are bolded/underlined.}
\label{tab:intervention}
\end{table*}

\begin{table*}[t]
\centering
\normalsize
\setlength{\tabcolsep}{8pt}
\begin{adjustbox}{max width=\textwidth}
\begin{tabular}{@{}lccccc@{}}
\toprule
Method & Loc. & Units & Sent. & Obs. & Pos.\\
\midrule
ChEX & $87.0\to87$ & $20.6\to25$ & $6.7\to6$ & $4.6\to3$ & $2.9\to1$\\
RGRG & $24.6\to10$ & $12.8\to10$ & $7.8\to3$ & $5.1\to3$ & $1.5\to0$\\
MAIRA-2 (grounded) & $1.7\to1$ & --- & $5.5\to6$ & $2.8\to2$ & $0.9\to0$\\
CURE & $31.7\to29$ & $29.0\to29$ & $41.4\to36$ & $8.1\to7$ & $4.3\to3$\\
PriorRG & --- & --- & $3.7\to3$ & $3.5\to2$ & $2.2\to1$\\
R2Gen-Mamba & --- & --- & $4.6\to4$ & $3.6\to4$ & $0.6\to1$\\
\midrule
RadCarve & $24.8\to6$ & $5.6\to1$ & $7.0\to1$ & $4.9\to0$ & $1.6\to0$\\
\bottomrule
\end{tabular}
\end{adjustbox}
\caption{Output counts: mean over 500 real images $\to$ all-black input. Loc.: localization outputs; Units: generation units before text postprocessing; Sent.: final sentences; Obs.: observed CheXbert categories; Pos.: positive or uncertain categories. Dashes indicate inapplicable counts.}
\label{tab:null_input}
\end{table*}

\subsection{Qualitative Examples}
\label{app:qualitative}
Figure~\ref{fig:qualitative} shows how RadCarve organizes clinically relevant descriptions through predicted region sets. In the upper example, bilateral lung and lower-lung-zone units capture pleural effusions and atelectasis, while a single-region cardiac unit describes cardiomegaly, covering key findings in the reference report at different anatomical granularities. However, the cardiac unit describes cardiomegaly as moderate, whereas the reference report describes mild enlargement. In the lower example, the bilateral lung unit describes clear lungs and the absence of pleural effusion or pneumothorax, while the hilar--mediastinal--cardiac unit captures normal contours and pulmonary vasculature. These examples illustrate how single- and multi-region units capture relevant report content with explicit links to regional inputs.

\subsection{Evidence Composition Intervention}
\label{app:intervention_protocol}
Of the 114,403 multi-region pairs in the given-set evaluation, 86,230 have all target regions correctly localized by RadCarve at $\mathrm{IoU}\geq0.5$. After excluding 219 pairs with insufficient eligible random controls, the intervention evaluates 86,011 pairs from 30,812 images. The image, generator, target text, and pre-extracted regional features are held fixed, and neither the planner nor its region-set token is used. Only the input region set changes. Best singleton selects the target region that maximizes target-text likelihood. Leave-one-out removes one target region at a time and averages over removals. Random same-size averages over five fixed equal-cardinality sets sampled per pair from correctly localized nontarget regions in the same image. Regions related to a target region by a direct anatomical redundancy rule are excluded from this sampling pool.

Each condition is scored by mean per-token negative log-likelihood under teacher forcing. For each pair, $\Delta$LL is the per-token log-likelihood difference between Full set and a control. Table~\ref{tab:intervention} gives the numerical results underlying the distributions in Figure~\ref{fig:intervention}. Confidence intervals for mean paired differences use 10,000 paired bootstrap resamples of the 30,812 images (Appendix~\ref{app:statistical_analysis}).

\subsection{Report-Level Metrics}
All complete-report methods are scored on the same images and reference reports after method-specific output processing removes non-report tokens, including serialized coordinates. A common normalization of whitespace and punctuation precedes scoring. BLEU \citep{papineni2002bleu} and METEOR \citep{banerjee2005meteor} are aggregated at the corpus level; ROUGE-L \citep{lin2004rouge} and CXRFES are averaged over reports. CXRFES extracts radiology facts and compares CXRFE-encoded facts in both matching directions \citep{messina2024cxrfe}. For complete-report test evaluation, candidate and reference reports are separately split into sentences, labeled with the same 14-label CheXbert model \citep{smit2020chexbert}, and merged into report-level binary labels. Following RGRG's label-state mapping \citep{tanida2023rgrg}, the micro metrics treat positive and uncertain states as positive, whereas example-based F1 uses positive states only. Following CURE's sentence-level label aggregation \citep{messina2026cure}, for each metric, the first 13 labels are merged by logical OR across sentences, and No Finding by logical AND. F1$_{\mathrm{mic}\text{-}5}$ uses cardiomegaly, edema, consolidation, atelectasis, and pleural effusion; F1$_{\mathrm{micro}\text{-}14}$ pools all 14 labels, while F1$_{\mathrm{ex}\text{-}14}$ averages per-report F1 over those labels.

\subsection{All-Black Input}
\label{app:null_input}
We use an all-black input as a qualitative example of the methods' end-to-end generation behavior and compare their output counts with averages on real images (Table~\ref{tab:null_input}).

\paragraph{Inputs and configurations.}
The real-image reference in Table~\ref{tab:null_input} uses the same 500 test images for all methods, sampled uniformly with seed 42 from images eligible for given-set evaluation. The black input has zero-valued pixels before each method's native preprocessing. Models receive no target region sets, reference text, prior examinations, or clinical context. ChEX uses its official report-generation configuration with 15 anatomical and 14 pathology queries, including its native filtering and postprocessing \citep{muller2025chex}. CURE uses AGRG-29+GRG \citep{messina2026cure}, and MAIRA-2 uses grounded report generation \citep{bannur2024maira2}. RGRG \citep{tanida2023rgrg} and RadCarve run their complete detection, unit-selection, and generation pipelines. Black-image entries report outputs for a fixed all-black input.

\paragraph{Counting rules.}
Loc. counts native localization outputs, not correctly localized regions: raw boxes before filtering for ChEX, detected regions for RGRG and RadCarve, boxes generated in MAIRA-2 grounded reports, and the total boxes generated by CURE’s AGRG and GRG branches. ChEX produces three boxes per query, hence 87 raw boxes. Units counts the visual conditioning units passed to text generation before text postprocessing; for CURE, it counts the 29 AGRG queries only. Sent. counts nonempty sentences in the final processed report using spaCy's sentencizer. CheXbert counts cover its 12 observation categories, excluding Support Devices and No Finding. For both real-image and black-input outputs, CheXbert is applied to NLTK-split sentences. Obs. counts categories labeled positive, negative, or uncertain in any sentence, while Pos. counts categories labeled positive or uncertain in any sentence. Each count takes the union across sentences, counting each category at most once per report. These are category counts per report, not counts of individual claims.

\paragraph{Output behavior.}
Relative to their real-image averages, the methods show different changes in unit output on the all-black input (Table~\ref{tab:null_input}). ChEX generates 25 units, exceeding its real-image average of 20.6, while CURE generates all 29 queried units under both conditions. RGRG generates ten units compared with a real-image average of 12.8. RadCarve's output contracts from an average of 5.6 units on real images to one unit on the black input. All tested baselines still produce text with CheXbert observation labels, whereas RadCarve's remaining sentence receives none.

The detection and selection outputs provide a complementary view of this behavior. RGRG's region selector retains all ten detected regions for generation. RadCarve's detector still outputs six regions, but its planner predicts only one generation unit containing a single region, the left lung. Together with the change relative to its real-image output, this example illustrates how the planner adjusts unit generation in response to image content despite persistent detector responses.

\paragraph{Example outputs.}
ChEX generates ``The heart is mildly enlarged.'' CURE generates ``There is a new left lower lobe opacity, which may represent pneumonia.'' RadCarve retains the singleton region set \{left lung\} and generates ``In comparison with the study of \_\_\_, there is little overall change in the appearance of the heart and lungs.'' Although this last output receives no CheXbert observation labels, its comparison is also unsupported by the blank input.

\section{Ablation and Sensitivity Analyses}
\label{app:ablation_diagnostics}
\label{app:ablation_protocol}

\begin{table*}[!t]
\centering
\normalsize
\setlength{\tabcolsep}{8pt}
\begin{tabular}{@{}lccccc@{}}
\toprule
Method & \metrichead{P$_{\mathrm{micro}\text{-}14}$} & \metrichead{R$_{\mathrm{micro}\text{-}14}$} & Units / report & Duplicate rate\,$\downarrow$ & Final length / Ref.\\
\midrule
RadCarve & \textbf{0.564} & \textbf{0.507} & 5.60 & \textbf{0.139} & 0.981\\
w/o Set Binding & 0.556 & 0.393 & 9.30 & 0.310 & 0.801\\
\bottomrule
\end{tabular}
\caption{Clinical content recovery and generation statistics after singleton decomposition. Duplicate rate measures exact sentence repetitions before deduplication; Final length / Ref.\ is the final-to-reference report word-count ratio. Best results are bolded.}
\label{tab:singleton}
\end{table*}

\subsection{Generation-Unit Construction Ablations}
\label{app:binding_ablation}
\paragraph{Region-set binding.}
For w/o Set Binding, the detector and planner are retained. We collect regions from all predicted sets, deduplicate by anatomical identity, and independently generate text for each resulting region. The generator is retrained using the corresponding single-region supervision. Thus, the anatomical regions covered by the report remain unchanged, but the grouping of regions into generation units is removed. The additional diagnostics in Table~\ref{tab:singleton} show a substantial drop in clinical recall with a smaller decrease in precision, along with more repeated sentences and shorter final reports.

\paragraph{Redundancy pruning and Hybrid.}
\label{app:pruning_ablation}
For w/o Pruning, pruning is disabled throughout training and inference, and both the planner and generator are retrained using raw region sets and their corresponding text targets. Hybrid retains this unpruned model's planner outputs, raw region sets, and region-set tokens, replacing only its generator with the full RadCarve generator trained under pruned supervision. Inference-time pruning remains disabled, so Hybrid receives the same regional inputs as the unpruned model.

Without pruning, 44.17\% of predicted sets retain a parent--child redundancy. Hybrid recovers much of the clinical-score gap while receiving these same unpruned sets, indicating that the supervision constructed after pruning contributes substantially to the improvement.

\begin{table*}[!t]
\begingroup
\centering
\normalsize
\setlength{\tabcolsep}{6pt}
\begin{adjustbox}{max width=\linewidth}
\begin{tabular}{@{}lccccccc@{}}
\toprule
Method & \metrichead{Set IoU} & \metrichead{P@0.5} & \metrichead{R@0.5} & \metrichead{F1@0.5} & Pred/GT & \metrichead{Single IoU} & \metrichead{Multi IoU}\\
\midrule
RGRG & \underline{0.471} & \underline{0.231} & \underline{0.613} & \underline{0.336} & 2.648 & \textbf{0.729} & \underline{0.378}\\
ChEX & 0.156 & 0.149 & 0.192 & 0.168 & 1.290 & 0.285 & 0.109\\
RadCarve & \textbf{0.567} & \textbf{0.551} & \textbf{0.642} & \textbf{0.593} & 1.165 & \underline{0.415} & \textbf{0.622}\\
\bottomrule
\end{tabular}
\end{adjustbox}
\caption{Region-set recovery under region-only matching, using the same test set and metrics as Table~\ref{tab:planner_ablations}. Best/second-best results are bolded/underlined.}
\label{tab:region_only_comparison}
\endgroup
\par\medskip
\begingroup
\centering
\normalsize
\setlength{\tabcolsep}{6pt}
\begin{adjustbox}{max width=\linewidth}
\begin{tabular}{@{}lcccccc@{}}
\toprule
Decoder layer & 1 & 2 & 3 & 4 & 5 & 6\\
\midrule
RadCarve & \textbf{0.563} & \textbf{0.564} & \textbf{0.566} & \textbf{0.566} & \textbf{0.567} & \textbf{0.567}\\
w/o Refinement & 0.541 & 0.546 & 0.549 & 0.550 & 0.551 & 0.553\\
\bottomrule
\end{tabular}
\end{adjustbox}
\caption{Set IoU ($\uparrow$) across all six decoder layers under the protocol of Table~\ref{tab:planner_ablations}. Best results at each layer are bolded.}
\label{tab:refinement_layers}
\endgroup
\par\medskip
\begingroup
\centering
\normalsize
\setlength{\tabcolsep}{6pt}
\begin{adjustbox}{max width=\linewidth}
\begin{tabular}{@{}lrcccc@{}}
\toprule
 & & \multicolumn{2}{c}{RadCarve} & \multicolumn{2}{c}{w/o Specialized Banks}\\
\cmidrule(lr){3-4}\cmidrule(lr){5-6}
Target type & GT units & \metrichead{Set IoU} & \metrichead{F1@0.5} & \metrichead{Set IoU} & \metrichead{F1@0.5}\\
\midrule
Broad & 124,449 & \textbf{0.620} & \textbf{0.649} & 0.600 & 0.641\\
Cross & 16,737 & \textbf{0.342} & \textbf{0.335} & 0.270 & 0.271\\
Specific & 10,121 & \textbf{0.287} & \textbf{0.296} & 0.208 & 0.233\\
\bottomrule
\end{tabular}
\end{adjustbox}
\caption{Region-set recovery by target composition type. Overall region-only assignments are reused without rematching within types; Set IoU averages over each type's targets. Best results for each type and metric are bolded.}
\label{tab:types}
\endgroup
\end{table*}

\subsection{Planner Ablations and Recovery Diagnostics}
\label{app:planner_diagnostics}
\paragraph{w/o Iterative Composition Refinement.}
The planner and generator are retrained after removing composition feedback. Each decoder layer still predicts region composition and receives the same intermediate supervision, but the predicted composition is no longer used to aggregate regional context and feed it into the next layer's set-level representations.

\paragraph{w/o Composition-Specialized Query Banks.}
The Broad, Cross, and Specific query banks are replaced with a homogeneous pool of the same total size. Within-type matching and bank-specific prediction and constraint mechanisms are removed. All queries share prediction heads and a unified matching procedure, and both the planner and generator are retrained.

\begin{table*}[!t]
\begingroup
\centering
\normalsize
\setlength{\tabcolsep}{6pt}
\begin{adjustbox}{max width=\linewidth}
\begin{tabular}{@{}lcccc@{}}
\toprule
Bank & Unit share & p99 / report & Queries & Coverage\\
\midrule
Broad & 81.87\% & 8 & 8 & 99.66\%\\
Cross & 10.81\% & 3 & 2 & 98.39\%\\
Specific & 7.32\% & 2 & 2 & 99.48\%\\
\bottomrule
\end{tabular}
\end{adjustbox}
\caption{Training-unit distribution and query allocation by composition type. p99 denotes the 99th percentile of units per report; Coverage is the fraction of reports whose target count does not exceed the bank's query capacity.}
\label{tab:bank_allocation}
\endgroup
\par\medskip
\begingroup
\centering
\normalsize
\setlength{\tabcolsep}{7pt}
\begin{tabular}{@{}lccccc@{}}
\toprule
Broad/Cross/Specific & \metrichead{Set IoU} & \metrichead{P@0.5} & \metrichead{R@0.5} & \metrichead{F1@0.5} & Pred/GT\\
\midrule
$4/2/2$ & 0.455 & \textbf{0.615} & 0.524 & 0.566 & 0.853\\
$8/2/2$ (default) & 0.567 & 0.551 & 0.642 & \underline{0.593} & 1.165\\
$12/2/2$ & \textbf{0.579} & 0.535 & \textbf{0.655} & 0.589 & 1.225\\
$10/1/1$ & \underline{0.573} & 0.549 & \underline{0.648} & \textbf{0.595} & 1.180\\
$6/3/3$ & 0.506 & \underline{0.590} & 0.577 & 0.583 & 0.979\\
\bottomrule
\end{tabular}
\caption{Capacity and allocation sensitivity of separately trained planners under the protocol of Table~\ref{tab:planner_ablations}. Best/second-best results are bolded/underlined.}
\label{tab:bank_sensitivity}
\endgroup
\par\medskip
\begingroup
\centering
\normalsize
\setlength{\tabcolsep}{5pt}
\begin{tabular}{@{}lcccccc@{}}
\toprule
& \multicolumn{2}{c}{Broad} & \multicolumn{2}{c}{Cross} & \multicolumn{2}{c}{Specific}\\
\cmidrule(lr){2-3}\cmidrule(lr){4-5}\cmidrule(lr){6-7}
Allocation & \metrichead{F1@0.5} & Pred/GT & \metrichead{F1@0.5} & Pred/GT & \metrichead{F1@0.5} & Pred/GT\\
\midrule
$4/2/2$ & 0.618 & 0.831 & \underline{0.136} & 0.602 & 0.099 & 1.536\\
$8/2/2$ & \textbf{0.640} & 1.188 & 0.134 & 0.627 & \underline{0.105} & 1.774\\
$12/2/2$ & 0.634 & 1.257 & \textbf{0.139} & 0.662 & \textbf{0.109} & 1.754\\
$10/1/1$ & \underline{0.637} & 1.239 & 0.000 & 0.000 & 0.073 & 2.409\\
$6/3/3$ & 0.618 & 1.058 & 0.068 & 0.356 & 0.099 & 1.034\\
\bottomrule
\end{tabular}
\caption{Per-bank recovery for the capacity configurations. Matching is performed separately within each bank, unlike the overall assignments reused in Table~\ref{tab:types}. Best/second-best F1 results are bolded/underlined.}
\label{tab:bank_activity}
\endgroup
\end{table*}

\begin{table*}[!t]
\centering
\normalsize
\setlength{\tabcolsep}{4pt}
\begin{adjustbox}{max width=\linewidth}
\begin{tabular}{@{}lcccccccc@{}}
\toprule
Method & \metrichead{B1} & \metrichead{B4} & \metrichead{MTR} & \metrichead{R-L} & \metrichead{CXRFES} & \metrichead{F1$_{\mathrm{mic}\text{-}5}$} & \metrichead{F1$_{\mathrm{micro}\text{-}14}$} & \metrichead{F1$_{\mathrm{ex}\text{-}14}$}\\
\midrule
RadCarve & \underline{0.390} & \underline{0.131} & \underline{0.169} & \textbf{0.268} & \underline{0.724} & \textbf{0.566} & \textbf{0.534} & \textbf{0.490}\\
w/o Region-Set Token & \textbf{0.391} & 0.130 & \textbf{0.170} & \underline{0.267} & \underline{0.724} & 0.549 & 0.520 & 0.470\\
w/o Local Rad-DINO Fusion & \underline{0.390} & \textbf{0.132} & \underline{0.169} & \textbf{0.268} & \textbf{0.725} & \underline{0.561} & \underline{0.523} & \underline{0.484}\\
\bottomrule
\end{tabular}
\end{adjustbox}
\caption{Complete-report generator component ablations on the Chest ImaGenome processed test set. Local Rad-DINO removal affects only the generator. Best/second-best results are bolded/underlined.}
\label{tab:generator_components}
\end{table*}

\paragraph{Region-set recovery.}
Under the region-only matching protocol in Appendix~\ref{app:binding_protocol}, RadCarve achieves F1@0.5 of 0.593, compared with 0.336 for RGRG and 0.168 for ChEX, and the highest overall and Multi IoU scores (Table~\ref{tab:region_only_comparison}). RGRG has higher Single IoU. Removing refinement reduces Set IoU and F1@0.5 despite similar predicted unit counts. Removing specialized banks also reduces both measures, with F1@0.5 falling from 0.593 to 0.577 (Table~\ref{tab:planner_ablations}).

Table~\ref{tab:refinement_layers} compares region-set recovery across all six decoder layers. RadCarve achieves higher Set IoU than w/o Refinement at every layer, indicating stronger recovery throughout the trained decoder.

Table~\ref{tab:types} stratifies recovery by target composition type, reusing the overall region-only assignments without rematching. For prediction counts, matched predictions inherit their target's type, and unmatched predictions are assigned to the type of their highest-IoU target in the report. Set IoU averages over each type's targets, with unmatched targets contributing zero; F1@0.5 uses matched-set $\mathrm{IoU}\geq0.5$. Removing specialized banks reduces F1@0.5 by 0.064 for Cross and 0.062 for Specific sets, compared with 0.008 for Broad sets. The larger drops for Cross and Specific sets indicate that the banks provide greater benefits for compositions involving the difficult-region subset.

\subsection{Bank Capacity and Allocation Sensitivity}
\label{app:bank_sensitivity}
\paragraph{Training-set basis for the default.}
Training reports contain a median of four target units and a 99th percentile of nine. Considering total unit counts alone, a capacity of 12 covers 99.96\% of reports, truncating only 132 of 758,717 source units. Broad accounts for 81.87\% of units and has a per-report 99th percentile of eight, motivating eight queries. Cross and Specific account for 10.81\% and 7.32\%; two queries each cover 98.39\% and 99.48\% of reports, preserving dedicated capacity for both less frequent types (Table~\ref{tab:bank_allocation}).

\paragraph{Capacity and allocation sensitivity.}
Table~\ref{tab:bank_sensitivity} compares separately trained planners across five capacity and allocation configurations on the same 31,271-image test set using region-only matching. We first vary Broad capacity from 4 to 8 to 12, keeping Cross and Specific at two each. With $4/2/2$, Pred/GT falls to 0.853 and recall to 0.524, despite higher precision. With $12/2/2$, they rise to 1.225 and 0.655, with higher Set IoU but lower F1 than $8/2/2$. At fixed total capacity 12, $10/1/1$ yields slightly higher F1 (0.595 versus 0.593), while $6/3/3$ raises precision and brings Pred/GT closer to one at lower Set IoU (0.506 versus 0.567). No configuration leads on every measure.

\paragraph{Bank-level behavior.}
Table~\ref{tab:bank_activity} evaluates bank activity, assigning predictions by query-bank origin and targets by composition type, then matching within banks. Unlike Table~\ref{tab:types}, it uses within-bank rather than overall matching, so scores are not directly comparable. Under $10/1/1$, Cross produces almost no valid units despite slightly higher overall F1. Under $6/3/3$, expanding Cross and Specific to three queries reduces both banks' output and F1, with a larger F1 drop for Cross. More queries therefore need not improve these less common compositions. Together, training coverage and sensitivity results support $8/2/2$ as a balanced allocation, not the best on every metric.

\subsection{Generator Component Ablations}
\label{app:generator_ablations}
For w/o Region-Set Token, we train a generator without the region-set token $u_t$ while retaining the predicted region sets and their member-region tokens; the planner is unchanged. For w/o Local Rad-DINO Fusion, the generator uses detector-derived region features without its local Rad-DINO feature branch, while the planner continues to use Rad-DINO. Table~\ref{tab:generator_components} shows modest changes in text-overlap scores and CXRFES, but lower clinical F1 in both variants.

\begin{table*}[!t]
\centering
\normalsize
\setlength{\tabcolsep}{6pt}
\begin{tabular}{@{}llrcr@{}}
\toprule
Comparison & Metric & Difference & 95\% CI & $p$\\
\midrule
\multicolumn{5}{@{}l}{Given anatomical region sets}\\
RadCarve $-$ RGRG & Region IoU & $+0.0014$ & $[+0.0007,+0.0021]$ & $<0.001$\\
 & Pair Pass & $+0.0053$ & $[+0.0039,+0.0067]$ & $<0.001$\\
 & CXRFES & $+0.0496$ & $[+0.0482,+0.0510]$ & $<0.001$\\
 & CXRFES$_{\geq2}$ & $+0.0337$ & $[+0.0322,+0.0353]$ & $<0.001$\\
 & F1$_{\mathrm{mic}\text{-}5}$ & $+0.0639$ & $[+0.0601,+0.0676]$ & $<0.001$\\
 & F1$_{\mathrm{micro}\text{-}14}$ & $+0.0558$ & $[+0.0539,+0.0576]$ & $<0.001$\\
 & F1$_{\mathrm{ex}\text{-}14}$ & $+0.0295$ & $[+0.0279,+0.0311]$ & $<0.001$\\
\addlinespace
RadCarve $-$ CURE & Region IoU & $+0.0847$ & $[+0.0837,+0.0857]$ & $<0.001$\\
 & Pair Pass & $+0.0445$ & $[+0.0424,+0.0467]$ & $<0.001$\\
 & CXRFES & $+0.0561$ & $[+0.0543,+0.0580]$ & $<0.001$\\
 & CXRFES$_{\geq2}$ & $+0.0498$ & $[+0.0477,+0.0519]$ & $<0.001$\\
 & F1$_{\mathrm{mic}\text{-}5}$ & $+0.0457$ & $[+0.0410,+0.0505]$ & $<0.001$\\
 & F1$_{\mathrm{micro}\text{-}14}$ & $+0.0395$ & $[+0.0370,+0.0419]$ & $<0.001$\\
 & F1$_{\mathrm{ex}\text{-}14}$ & $+0.0448$ & $[+0.0426,+0.0469]$ & $<0.001$\\
\midrule
\multicolumn{5}{@{}l}{End-to-end binding after text-only matching}\\
RadCarve $-$ RGRG & Binding IoU & $+0.1624$ & $[+0.1598,+0.1650]$ & $<0.001$\\
 & F1@0.5 & $+0.2664$ & $[+0.2637,+0.2690]$ & $<0.001$\\
 & Single IoU & $+0.00499$ & $[-0.00003,+0.00986]$ & 0.0475\\
 & Multi IoU & $+0.2194$ & $[+0.2166,+0.2223]$ & $<0.001$\\
\addlinespace
RadCarve $-$ ChEX & Binding IoU & $+0.3442$ & $[+0.3410,+0.3474]$ & $<0.001$\\
 & F1@0.5 & $+0.3625$ & $[+0.3594,+0.3655]$ & $<0.001$\\
 & Single IoU & $+0.18787$ & $[+0.18313,+0.19253]$ & $<0.001$\\
 & Multi IoU & $+0.4008$ & $[+0.3972,+0.4044]$ & $<0.001$\\
\bottomrule
\end{tabular}
\caption{Selected paired differences for generation given region sets (Table~\ref{tab:given_sets}) and end-to-end binding recovery (Table~\ref{tab:binding}).}
\label{tab:stats_units}
\end{table*}

\section{Statistical Analysis}
\label{app:statistical_analysis}

\subsection{Paired Bootstrap Protocol}
We use 10,000 paired nonparametric bootstrap resamples, holding predictions and references fixed. The resampling unit is an image/report; all text--region-set pairs from an image are sampled together, and compared methods use the same sampled indices. Each resample recomputes metrics with their original aggregation rules, including corpus-level text scores and micro-averaged clinical scores. Point estimates are computed on the original test set. Differences and relative improvements are computed from unrounded scores, and 95\% confidence intervals (CIs) for paired differences are the 2.5th and 97.5th bootstrap percentiles.

For an observed paired difference $\widehat\Delta$ and bootstrap differences $\Delta_b^*$, we use the two-sided centered-bootstrap test
\begin{equation}
p=\frac{1+\sum_{b=1}^{B}\mathbf{1}\!\left\{\left|\Delta_b^*-\widehat\Delta\right|\geq\left|\widehat\Delta\right|\right\}}{B+1}.
\end{equation}
Here $B=10{,}000$. We report the resulting p-values, using a nominal significance level of 0.05. Table~\ref{tab:intervention} already reports the paired differences and CIs for the evidence-composition intervention.

\begin{table*}[!t]
\begingroup
\centering
\normalsize
\setlength{\tabcolsep}{5pt}
\begin{tabular}{@{}llrcr@{}}
\toprule
Baseline ($\mathrm{RadCarve}-$) & Metric & Difference & 95\% CI & $p$\\
\midrule
\multicolumn{5}{@{}l}{In-domain report generation}\\
RGRG & B1 & $+0.0137$ & $[+0.0127,+0.0148]$ & $<0.001$\\
RGRG & B4 & $+0.0044$ & $[+0.0034,+0.0053]$ & $<0.001$\\
RGRG & METEOR & $+0.0026$ & $[+0.0020,+0.0031]$ & $<0.001$\\
R2Gen-Mamba & R-L & $-0.00003$ & $[-0.0013,+0.0012]$ & 0.962\\
RGRG & CXRFES & $+0.0038$ & $[+0.0031,+0.0045]$ & $<0.001$\\
MAIRA-2 (non-grounded) & F1$_{\mathrm{mic}\text{-}5}$ & $+0.0144$ & $[+0.0098,+0.0190]$ & $<0.001$\\
MAIRA-2 (non-grounded) & F1$_{\mathrm{micro}\text{-}14}$ & $+0.0079$ & $[+0.0046,+0.0111]$ & $<0.001$\\
MAIRA-2 (non-grounded) & F1$_{\mathrm{ex}\text{-}14}$ & $+0.0072$ & $[+0.0035,+0.0109]$ & $<0.001$\\
\midrule
\multicolumn{5}{@{}l}{External report generation on PadChest-GR}\\
PriorRG & CXRFES & $-0.0066$ & $[-0.0149,+0.0012]$ & 0.105\\
RGRG & F1$_{\mathrm{mic}\text{-}5}$ & $+0.0881$ & $[+0.0456,+0.1293]$ & $<0.001$\\
R2Gen-Mamba & F1$_{\mathrm{micro}\text{-}14}$ & $+0.0525$ & $[+0.0237,+0.0808]$ & $<0.001$\\
R2Gen-Mamba & F1$_{\mathrm{ex}\text{-}14}$ & $+0.0197$ & $[-0.0038,+0.0427]$ & 0.095\\
\bottomrule
\end{tabular}
\caption{Paired differences against the highest-scoring baseline for each of the eight in-domain metrics (Table~\ref{tab:indomain}) and the four external metrics (Table~\ref{tab:external}).}
\label{tab:stats_reports}
\endgroup
\end{table*}

\subsection{Generation Units and Evidence Binding}
Table~\ref{tab:stats_units} reports paired differences against RGRG and CURE for generation given region sets, and against RGRG and ChEX for end-to-end binding recovery. The small localization differences against RGRG contrast with larger text-score differences, while the end-to-end results show gains in overall binding recovery and multi-region IoU. The Single IoU difference against RGRG is small and borderline: its percentile CI slightly crosses zero, whereas the centered-bootstrap p-value is 0.0475. We therefore do not claim a clear advantage on single-region recovery.

\subsection{Complete-Report Generation}
Table~\ref{tab:stats_reports} compares RadCarve with the highest-scoring baseline for each metric in the corresponding main table. The in-domain comparison includes all eight reported report-quality metrics; the external comparison includes CXRFES and all three clinical metrics. On PadChest-GR, the gains in the two micro-F1 metrics are significant, whereas the differences in CXRFES against PriorRG and F1$_{\mathrm{ex}\text{-}14}$ against R2Gen-Mamba are not statistically significant.

\begin{table*}[!t]
\centering
\normalsize
\setlength{\tabcolsep}{5pt}
\begin{adjustbox}{max width=\linewidth}
\begin{tabular}{@{}llrcr@{}}
\toprule
Comparison & Metric & Difference & 95\% CI & $p$\\
\midrule
\multicolumn{5}{@{}l}{Main ablations}\\
RadCarve $-$ w/o Set Binding & F1$_{\mathrm{mic}\text{-}5}$ & $+0.0585$ & $[+0.0547,+0.0622]$ & $<0.001$\\
 & F1$_{\mathrm{micro}\text{-}14}$ & $+0.0731$ & $[+0.0704,+0.0758]$ & $<0.001$\\
 & F1$_{\mathrm{ex}\text{-}14}$ & $+0.0521$ & $[+0.0494,+0.0549]$ & $<0.001$\\
\addlinespace
RadCarve $-$ w/o Pruning & F1$_{\mathrm{mic}\text{-}5}$ & $+0.0544$ & $[+0.0499,+0.0590]$ & $<0.001$\\
 & F1$_{\mathrm{micro}\text{-}14}$ & $+0.0520$ & $[+0.0489,+0.0551]$ & $<0.001$\\
 & F1$_{\mathrm{ex}\text{-}14}$ & $+0.0501$ & $[+0.0468,+0.0533]$ & $<0.001$\\
\addlinespace
RadCarve $-$ w/o Refinement & F1$_{\mathrm{mic}\text{-}5}$ & $+0.0464$ & $[+0.0424,+0.0505]$ & $<0.001$\\
 & F1$_{\mathrm{micro}\text{-}14}$ & $+0.0538$ & $[+0.0510,+0.0567]$ & $<0.001$\\
 & F1$_{\mathrm{ex}\text{-}14}$ & $+0.0447$ & $[+0.0417,+0.0477]$ & $<0.001$\\
\addlinespace
RadCarve $-$ w/o Specialized Banks & F1$_{\mathrm{mic}\text{-}5}$ & $+0.0274$ & $[+0.0235,+0.0313]$ & $<0.001$\\
 & F1$_{\mathrm{micro}\text{-}14}$ & $+0.0353$ & $[+0.0326,+0.0380]$ & $<0.001$\\
 & F1$_{\mathrm{ex}\text{-}14}$ & $+0.0295$ & $[+0.0266,+0.0323]$ & $<0.001$\\
\midrule
\multicolumn{5}{@{}l}{Hybrid comparisons}\\
RadCarve $-$ Hybrid & F1$_{\mathrm{mic}\text{-}5}$ & $+0.0070$ & $[+0.0031,+0.0107]$ & $<0.001$\\
 & F1$_{\mathrm{micro}\text{-}14}$ & $+0.0100$ & $[+0.0073,+0.0127]$ & $<0.001$\\
 & F1$_{\mathrm{ex}\text{-}14}$ & $+0.0175$ & $[+0.0147,+0.0203]$ & $<0.001$\\
\addlinespace
Hybrid $-$ w/o Pruning & F1$_{\mathrm{mic}\text{-}5}$ & $+0.0474$ & $[+0.0430,+0.0517]$ & $<0.001$\\
 & F1$_{\mathrm{micro}\text{-}14}$ & $+0.0419$ & $[+0.0390,+0.0448]$ & $<0.001$\\
 & F1$_{\mathrm{ex}\text{-}14}$ & $+0.0325$ & $[+0.0296,+0.0355]$ & $<0.001$\\
\midrule
\multicolumn{5}{@{}l}{Generator components}\\
RadCarve $-$ w/o Region-Set Token & F1$_{\mathrm{mic}\text{-}5}$ & $+0.0173$ & $[+0.0143,+0.0205]$ & $<0.001$\\
 & F1$_{\mathrm{micro}\text{-}14}$ & $+0.0137$ & $[+0.0116,+0.0159]$ & $<0.001$\\
 & F1$_{\mathrm{ex}\text{-}14}$ & $+0.0205$ & $[+0.0182,+0.0228]$ & $<0.001$\\
\addlinespace
RadCarve $-$ w/o Local Rad-DINO Fusion & F1$_{\mathrm{mic}\text{-}5}$ & $+0.0053$ & $[+0.0024,+0.0083]$ & $<0.001$\\
 & F1$_{\mathrm{micro}\text{-}14}$ & $+0.0103$ & $[+0.0083,+0.0123]$ & $<0.001$\\
 & F1$_{\mathrm{ex}\text{-}14}$ & $+0.0069$ & $[+0.0048,+0.0089]$ & $<0.001$\\
\bottomrule
\end{tabular}
\end{adjustbox}
\caption{Paired differences on all three clinical F1 metrics for the main complete-report ablations and Hybrid (Table~\ref{tab:ablations}), and generator component ablations (Table~\ref{tab:generator_components}). All metrics are higher-is-better.}
\label{tab:stats_ablation_clinical}
\end{table*}

\newpage
\subsection{Key Ablation Comparisons}
Table~\ref{tab:stats_ablation_clinical} reports all three clinical F1 differences for the main and generator component ablations. Hybrid is compared with w/o Pruning to quantify the recovery obtained by changing the generator while keeping its unpruned inputs, and with full RadCarve to measure the remaining gap. All listed clinical differences have positive 95\% CIs and $p<0.001$.

Table~\ref{tab:stats_ablations} reports singleton and planner diagnostics. The clinical-recall and duplicate-rate rows characterize the effect of singleton decomposition, while region-only matching evaluates planner recovery without generated text. Composition-type comparisons show positive IoU and F1@0.5 differences for Broad, Cross, and Specific sets. These tests assess the full model against w/o Specialized Banks within each type, rather than differences in improvement between types.

\begin{table*}[!t]
\centering
\normalsize
\setlength{\tabcolsep}{5pt}
\begin{adjustbox}{max width=\linewidth}
\begin{tabular}{@{}llrcr@{}}
\toprule
Comparison & Metric & Difference & 95\% CI & $p$\\
\midrule
\multicolumn{5}{@{}l}{Singleton and overall planner diagnostics}\\
RadCarve $-$ w/o Set Binding & R$_{\mathrm{micro}\text{-}14}$ & $+0.1136$ & $[+0.1107,+0.1166]$ & $<0.001$\\
 & Duplicate rate & $-0.1705$ & $[-0.1721,-0.1689]$ & $<0.001$\\
\addlinespace
RadCarve $-$ w/o Refinement & Set IoU & $+0.0138$ & $[+0.0126,+0.0150]$ & $<0.001$\\
\addlinespace
RadCarve $-$ w/o Specialized Banks & F1@0.5 & $+0.0158$ & $[+0.0144,+0.0172]$ & $<0.001$\\
\midrule
\multicolumn{5}{@{}l}{Recovery by target composition type}\\
RadCarve $-$ w/o Specialized Banks & Broad IoU & $+0.0193$ & $[+0.0180,+0.0206]$ & $<0.001$\\
 & Broad F1@0.5 & $+0.0081$ & $[+0.0066,+0.0095]$ & $<0.001$\\
\addlinespace
 & Cross IoU & $+0.0712$ & $[+0.0659,+0.0764]$ & $<0.001$\\
 & Cross F1@0.5 & $+0.0640$ & $[+0.0574,+0.0704]$ & $<0.001$\\
\addlinespace
 & Specific IoU & $+0.0789$ & $[+0.0720,+0.0858]$ & $<0.001$\\
 & Specific F1@0.5 & $+0.0623$ & $[+0.0549,+0.0698]$ & $<0.001$\\
\bottomrule
\end{tabular}
\end{adjustbox}
\caption{Paired differences for singleton diagnostics (Table~\ref{tab:singleton}), overall planner recovery (Table~\ref{tab:planner_ablations}), and recovery by composition type (Table~\ref{tab:types}). Duplicate rate is lower-is-better; the other metrics are higher-is-better.}
\label{tab:stats_ablations}
\end{table*}

\end{document}